\documentclass[10pt,journal,compsoc]{IEEEtran}
\ifCLASSOPTIONcompsoc
  \usepackage[noadjust]{cite}
\else
  \usepackage[noadjust]{cite}
\fi
\ifCLASSINFOpdf
\else
\fi
\usepackage{amsmath,amsfonts,amssymb,amsthm}
\usepackage{mathtools}
\usepackage{commath}
\usepackage[sc,osf]{mathpazo}

\usepackage{bbm}
\usepackage{amsfonts}
\usepackage{graphicx}
\usepackage{caption}
\usepackage{subcaption}
\usepackage{authblk}

\usepackage{textcomp}

\usepackage{algorithmic}

\usepackage[]{algorithm2e}
\usepackage{pgfplots}
\usepackage{tikzscale}
\usepackage{cleveref}
\usepackage{tikzscale}
\begin{document}
%
\title{Multiplierless and Sparse Machine Learning based on Margin Propagation Networks}
\author[$\ast$]{{Nazreen P.M.}}
\author[$\dagger$]{{Shantanu Chakrabartty}}
\author[$\ast$]{{Chetan Singh Thakur}
}
\affil[$\ast$]{Department of Electronic Systems Engineering, Indian Institute of Science, Bangalore 560012\\{\small \tt nazreen.pm@gmail.com,csthakur@iisc.ac.in}}
\affil[$\dagger$]{Department of Electrical and Systems Engineering, Washington University in St. Louis,USA, 63130\\
	\small \tt shantanu@wustl.edu}
\IEEEtitleabstractindextext{%
\begin{abstract}

The new generation of machine learning processors have evolved from multi-core and parallel architectures (for example graphical processing units) that were designed to efficiently implement matrix-vector-multiplications (MVMs). This is because at the fundamental level, neural network and machine learning operations extensively use MVM operations and hardware compilers exploit the inherent parallelism in MVM operations to achieve hardware acceleration on GPUs, TPUs and FPGAs.
However, many IoT and edge computing platforms require embedded ML devices close to the network in order to compensate for communication cost and latency. Hence a natural question to ask is whether MVM operations are even necessary to implement ML algorithms and whether simpler hardware primitives can be used to implement an ultra-energy-efficient ML processor/architecture. In this paper we propose an alternate hardware-software codesign of ML and neural network architectures where instead of using MVM operations and non-linear activation functions, the architecture only uses simple addition and thresholding operations to implement inference and learning. At the core of the proposed approach is margin-propagation based computation that maps multiplications into additions and additions into a dynamic rectifying-linear-unit (ReLU) operations. This mapping results in significant improvement in computational and hence energy cost. The training of a margin-propagation (MP) network involves optimizing an $L_1$ cost function, which in conjunction with ReLU operations leads to network sparsity. In this paper, we show how the MP network formulation can be applied for designing linear classifiers, shallow multi-layer perceptrons and support vector networks suitable fot IoT platforms and tiny ML applications. We show that these MP based classifiers give comparable results to that of their traditional counterparts for benchmark UCI datasets, with the added advantage of reduction in computational complexity enabling an improvement in energy efficiency.

\end{abstract}

\begin{IEEEkeywords}
Margin Propagation, Low Power, Machine learning, Multi-layer Perceptron, Support Vector Machine, Approximate Computing, 
\end{IEEEkeywords}}

\maketitle

\IEEEdisplaynontitleabstractindextext

%
\IEEEpeerreviewmaketitle

\IEEEraisesectionheading{\section{Introduction}\label{sec:introduction}}

Reducing the energy footprint is one of the major goals in the design of current and future machine learning (ML) systems. This is not only applicable for deep-learning platforms that run on data servers, consuming mega-watts of power \cite{al2015efficient}, but is also applicable for Internet-of-things (IoT) and edge computing platforms that are highly energy-constrained~\cite{li2018learning}. Computation in most of these ML systems are highly regular and involve repeated use of matrix-vector-multiplication (MVM) and non-linear activation and pooling operations. Therefore, current hardware compilers achieve performance acceleration and energy-efficiency by optimizing these fundamental operations on parallel hardware like the Graphical Processing Units (GPUs) or the Tensor Processing Units (TPUs). This mapping onto hardware accelerators can be viewed as a top-down approach where the goal from the perspective of a hardware designers to efficiently but faithfully map well-established ML algorithms without modifying the basic MVM or the activation functions. However, many IoT and edge computing platforms require embedded ML devices and this calls for a reduction in the computational complexity of MVM operations using alternate, simpler operations.
If the MVMs and the non-linear activation-functions could be combined in a manner that the resulting architecture becomes multiplier-less and uses much simpler computational primitives, then significant energy-efficiency could be potentially achieved at the system-level. In this paper we argue that a margin-propagation (MP) based computation can achieve this simplification by mapping multiplications into additions and additions into a dynamic rectifying-linear-unit (ReLU) operations.


\begin{figure*}[h!]%
	
	
	\vspace{-1em}
	\centering

	\includegraphics[height=4cm]{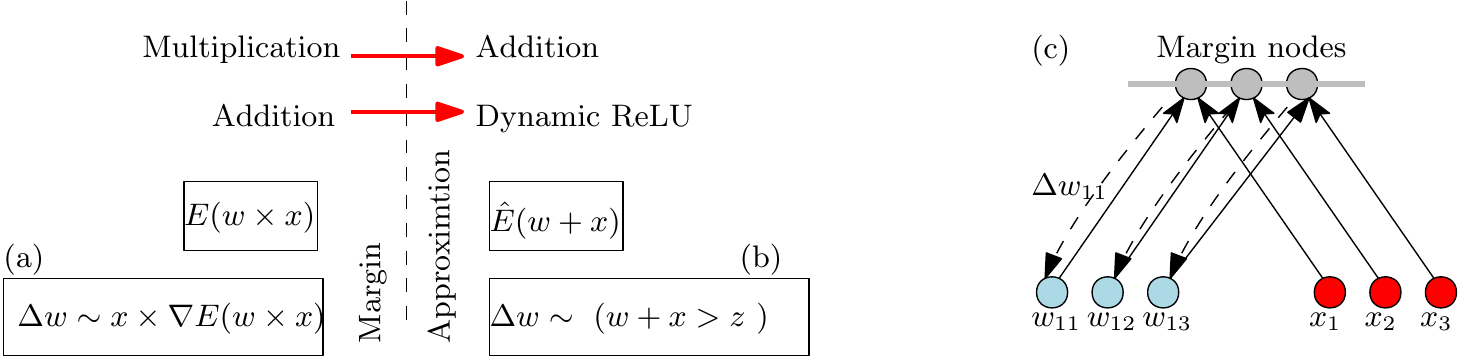}

	
	\caption{Hardware-software co-design using margin--propagation design framework to map multiplications into additions, and additions into dynamic rectifying linear operations: (a) Learning in the conventional architecture using a loss function E where parameter updates are estimated as the product of the gradient and the input; (b) Learning in the margin-propagation (MP) architecture where parameter updates are just Boolean up/down flag with no products; (c) Mapping of real-time learning architecture into margin-propagation architecture where parameter updates could be implemented using simple feedback paths. 
		}
	\label{fig_into}	
	
\end{figure*} 

The consequence of this mapping is a significant reduction in the complexity of inference and training which in turn leads to significant improvement in system energy-efficiency. To illustrate this, consider a very simple example as shown in Fig.\ref{fig_into}(a) and (b) for a comprising of a single training parameter w and a one-dimensional input x. In a conventional architecture minimizing a loss-function E(.) in Fig. \ref{fig_into}(a) results in a learning/parameter update step that requires modulating the gradient with the input. In the equivalent margin-approximation, as shown in Fig. \ref{fig_into}(b), the absence of multiplication implies that each parameter update is independent and the use of ReLU operations leads to learning update that involves only Boolean predicates. Rather than modulating the gradient with the input (as shown in Fig. \ref{fig_into}(a)), the new updates are based on comparing the sum of w and x with respect to a dynamic threshold z, as shown in Fig. \ref{fig_into}(b). This significantly simplifies the learning phase, and the storage of the parameters w. This is illustrated in Fig. \ref{fig_into}(c) using a single-layer network with three-dimensional input/parameters. The margin nodes not only implement the forward computation but also provide a continuous feedback to updates parameters $w_{11}-w_{13}$. For a digital implementation, this could be a simple up/down flag; for an analog implementation this could be equivalent to charging or discharging a capacitor storing the values of w11-w13.

Approximate computing research offers resource  savings at the cost of reduced accuracy and has gained much popularity as energy efficiency is much needed in several applications such as machine learning, signal processing, big data analytics, edge ML etc where certain amount of computational errors can be tolerated. Some of the notable works in the field includes approximate neural networks, approximate adders and video image processing, neural acceleration, energy efficient neuromorphic systems  etc ~\cite{peng2018axnet,miao2012modeling,sekanina2017approximate,esmaeilzadeh2012neural,venkataramani2014axnn}. 
Research is also done in reducing the complexity of multiplication operation to design an approximate multiplier ~\cite{imani2018rmac,hashemi2015drum,narayanamoorthy2014energy,kulkarni2011trading,imani2016acam}.  Approximation techniques to reduce the complexity of multiplication in neural networks are proposed in ~\cite{imani2018canna,lin2015neural,mrazek2016design}.

A very popular area employing approximate computing is in IoT platforms. IoT devices such as sensors, remote cameras etc. often require additional computing ML devices close to the networks where data is generated in order to reduce communication costs and latency and also to account for data privacy. But these devices  have small memories and limited computational powers. The multiplierless approach based on MP algorithm that we propose in this paper results in significant reduction in complexity and improvement in energy efficiency and  will be particularly  useful for such highly energy constrained tiny ML applications. We show that shallow MLPs and SVMs using MP formulation suitable for IoT platforms give comparable results to that of their traditional counterparts for benchmark UCI datasets  \cite{blake1998uci}, with the added advantage of reduction in computational complexity and improvement in energy efficiency.

Margin-propagation (MP) is originally proposed in~\cite{chakrabartty2004margin} and then was used in~\cite{gu2009sparse,gu2012theory} in the context of approximate computing and synthesis of piece-wise linear circuits. In~\cite{chakrabartty2007gini,chakrabartty2005sub,kucher2007energy,gu2009sparse,gu2012theory} the MP formulation is
used to synthesize ML algorithms, by replacing the MVM operation with simple addition and thresholding operations. However, in all the previous formulations, MP was to approximate log-sum-exp and any approximation error would propagate/accumulate as the size of the network increased. The formulation presented in this paper
views MP as an independent computational paradigm and the networks presented in this paper are trained using the exact form of the MP function. 

The paper is organized as follows: Section \ref{mp} discuss the margin propagation (MP) algorithm and compare its computational complexity with traditional MVM. Section \ref{perceptron} presents MP based perceptron  and its simulation results. Similarly sec. \ref{mlp} and \ref{svm} discuss MP based MLP and SVM respectively and their simulation results. We also evaluate the performances of MP based MLP and MP based SVM on different benchmark UCI datasets \cite{blake1998uci} and show that both MP MLP and MP SVM are able to give performances at par with that of conventional MLP and SVM with an added advantage of significant reduction in computational complexity and improvement in system energy. 
 Section \ref{conc} concludes the paper.

\begin{figure}[h!]%
	
	
	\vspace{-1em}
	\centering

	\includegraphics[height=5cm]{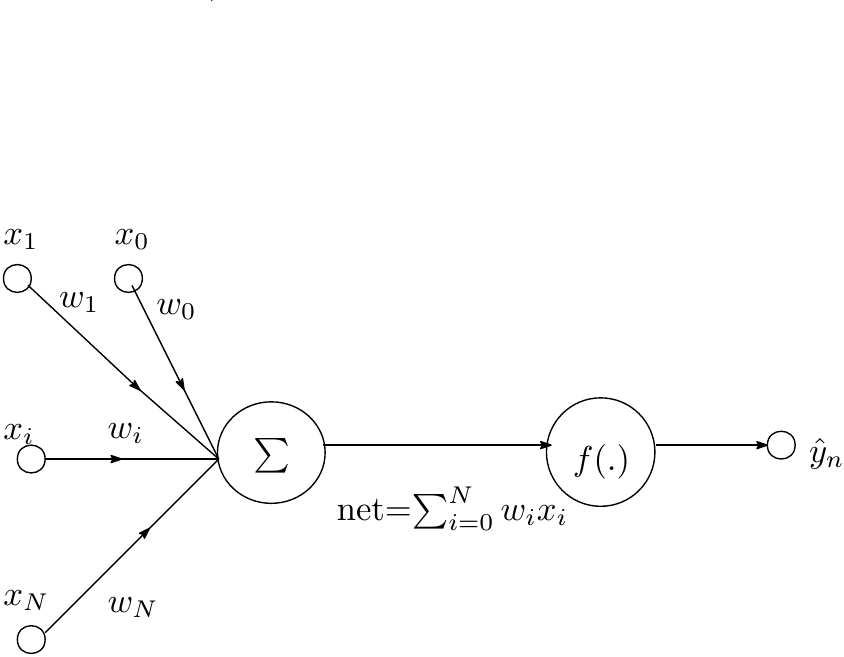}

	
	\caption{Perceptron as a binary classifier}
	\label{fig_perc}	
	
\end{figure}

A perceptron \cite{freund1999large,bishop2006pattern} is a single layer neural network used as a linear binary classifier as shown in fig. \ref{fig_perc}. Let input vector to a perceptron be $\bar{x}= \{x_i\}; 0\leq i\leq N$; where $x_0$ is the bias. The weighted sum of these inputs and the bias with the weights $\bar{w}= \{w_i\} ; 0\leq i \leq N$ is taken which is then fed into the activation function which maps the input into one of the two classes. For learning the perceptron weights standard gradient descent can be used with sum of squared errors as our cost function as given below;
\begin{equation}
E(\bar{w})= \frac{1}{2}\sum_{n}^{}[y_n-\hat{y}_n]^2
\label{eq_perc_cost}
\end{equation}  
where $y_n$ is the actual output for sample $n$ and $\hat{y}_n$ is the estimated output.

Support Vector Machine (SVM) is a supervised machine learning algorithm which is used mostly for classification problems \cite{cortes1995support}. Given labeled training data, SVM outputs an optimal hyperplane which categorizes any new test input into one of the classes. 
Given a test input $x_{i}; 1 \leq i \leq N$ where $x_{i} \in \mathbb{R}$, the decision function for SVM is given as,
\begin{equation}
f({x_{i}})= \sum_{s}^{}{w_s}{K}({x_s},{x_{i}})= \sum_{s}^{}({w_s}^+-{w_s}^-)({K_s}^+-{K_s}^-)
\label{smv_eq_intro}
\end{equation}
where ${K}$ is the kernel function, $x_s$ is the $s^{th}$ support vector and $x_i$ is the $i^{th}$ sample of the input vector.
\begin{figure}[h!]%
	
	
	\vspace{-1em}
	\centering

	\includegraphics[height=5cm]{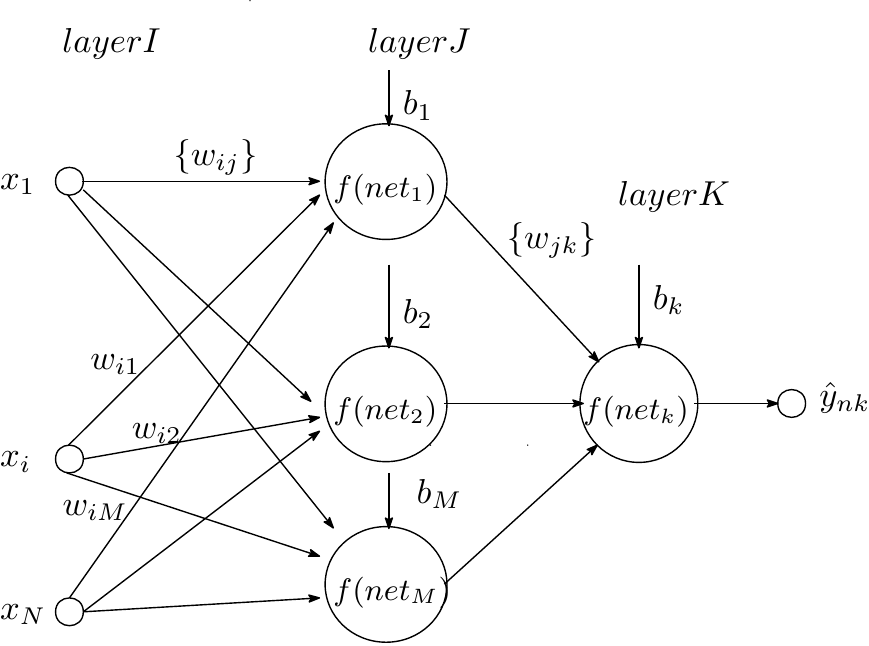}

	
	\caption{A three layer multilayer perceptron (MLP) for a two class problem; with $M$ hidden nodes and $N$ inputs.}
	\label{fig_mlp}	
	
\end{figure}   

In order to learn complex functions, a group of perceptrons can be stacked up in multiple layers  \cite{bishop2006pattern} to form a multilayer perceptron (MLP). A three layer MLP for a two class problem is shown in Fig. \ref{fig_mlp}. The weighted sum of the input vector $\bar{x}= \{x_i\}; 1\leq i\leq N$ with the weights $\bar{w}= \{w_{ij}\} ; 1\leq i \leq N; 1\leq j\leq M$ of the hidden layer is the input to the activation functions in the hidden layer. In the figure, $b_j, 1\leq j \leq M$ and $b_k, k=1$ indicates the input bias to each nodes in the hidden layer $J$ and output layer  $K$ whose weights are usually set to 1. Using the weights $w_{jk}; 1\leq j \leq M; k=1$ from the hidden layer to output layer, the weighted sum of the outputs from the hidden layer is again computed  which is then fed into the activation function of the final output node to obtain the output. The weights of such a feed-forward multilayer network is learned using the backpropagation algorithm. In this case also a squared error cost function is used.

\begin{equation}
E(\bar{w})= \frac{1}{2}\sum_{n}^{}\sum_{k}^{}[y_{nk}-\hat{y}_{nk}]^2
\label{eq_mlp_cost}
\end{equation}
  where $k \in output=1$, in this case.

\section{Margin propagation computation and complexity}\label{mp}
\begin{figure}[h!]%
	
	
	\vspace{-1em}
	\centering

	\includegraphics[height=4.7cm]{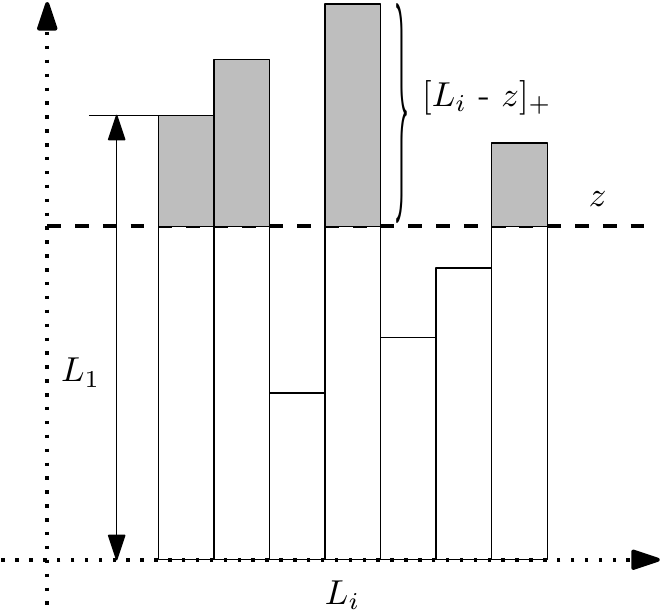}

	
	\caption{Reverse water-filling procedure}
	\label{fig_rwf}	
	
\end{figure}

MP algorithm is based on the reverse water filling procedure \cite{gu2009sparse,gu2012theory} as shown in Fig. \ref{fig_rwf}. The algorithm computes the normalization factor $z$, given a set of scores $L_i \in \mathbb{R}, 1 \leq i \leq N$ using the constraint;

\begin{equation}
\sum_{i=1}^{N}[L_i-z]_+ = \gamma
\end{equation}  
where $[.]_+=max(.,0)$ is the rectification operation and  $\gamma$ is the algorithm parameter.
This is a recursive algorithm which computes $z$ such that the net balance of score $L_i$ in excess to $z$ is $\gamma$ \cite{gu2009sparse,gu2012theory}. Thus given a set of input scores $L_i$, we can obtain the factor $z$ as;
\begin{equation}
z = \textit{MP}(\mathcal{L},\gamma)
\end{equation}
where $\mathcal{L} =\{L_i\}; 1\leq i\leq N$
\subsection{Complexity}\label{complexity1}
As mentioned before replacing the MVM operations in the perceptron, SVM and MLP into simple addition and thresholding operations in the log-likelihood domain using MP algorithm during inference and learning, significantly reduces the complexity. If $N$ is the  dimension of the input vector $\bar{x}$, then the overall complexity for an MVM operation,
\begin{equation}
z= \sum_{i}^{N}w_i x_i
\end{equation}
is 
\begin{align}
C_{MVM}= N \times C_M+ N \times C_A\\
\end{align}
where $C_{MVM}$ is the complexity of MVM operation. $C_M$ and $C_A$ are the complexity of multiplication and addition operations.

whereas for the margin propagation algorithm
\begin{equation}
z = \textit{MP}(\mathcal{L},\gamma)
\end{equation}
 the overall complexity is given as \cite{gu2012theory},
\begin{align}
C_{MP}=  N \times C_A + F \times log(N) \times C_c\\
\end{align}
where $F$ is the sparsity factor of the thresholding operation determined by $\gamma$ and $C_c$ is the complexity of comparison operation which is an elementary operation. 

In \cite{energypaper}, they show that for a digital circuit, if shift of one bit is defined as one elementary operation, the full adder requires about 3 operations per bit. Hence complexity of addition , $C_A= 3 \times C_S$ where $C_S$ is the complexity of shift operation. They also show that, 2 complete multiplications require $d^2$ full adders, where d is the number of bits.

Hence replacing MVM using MP algorithm will result in significant improvement in energy cost, as energy per multiplication is more than energy per addition operation as explored in \cite{horowitz20141}. In \cite{horowitz20141}, they show that for an 8 bit integer multiplication the rough energy cost is $0.2 pJ$ with a relative area cost of $282 \mu m^2$ whereas for an 8 bit addition  it is only $0.03 pJ$ and $36 \mu m^2$. For $32$ bit integer case, the energy cost is $3.1 pJ$ and area cost is $3495 \mu m^2$ for multiplication and $0.1 pJ$ and $137 \mu m^2$ for addition. The $L_1$ cost function used in conjunction with the ReLU operation ensures network sparsity.

\section{Perceptron using MP algorithm}\label{perceptron}
\begin{figure}[htbp]%
	
	
	\vspace{-1em}
	\centering

	\includegraphics[height=4.8cm]{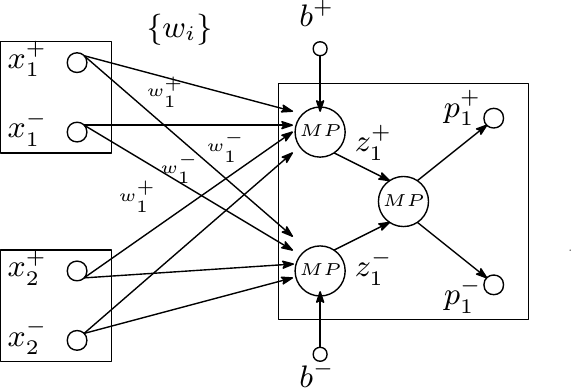}

	
	\caption{Perceptron using margin propagation (MP) algorithm, as a binary classifier for linearly separable data.}
	\label{fig_perc_mp}	
	
\end{figure}

A single layer perceptron using MP algorithm is shown in Fig. \ref{fig_perc_mp}. We minimize the $l_1$ norm given in  eq. \eqref{eq1_per} as the cost function to learn the network parameters. The inputs and weights are in the log-likelihood domain so that the network can be implemented using MP algorithm as mentioned in \cite{gu2012theory}.

\subsection{Inference}
Let the input vector to the perceptron in the log-likelihood domain be ${x}= \{x_i\}; 1\leq i\leq N$ and let $\{w_i\}$ be the learned weights.

From Fig \ref{fig_perc_mp} the perceptron output in differential form is, 
\begin{equation}
p(x) = p^+- p^-
\label{eq33_per}
\end{equation}

For the output node;
\begin{align}
p^+ & =[z^+-z]_+ \nonumber \\
p^- & =[z^--z]_+
\label{eq2_per}
\end{align}
where $z$ is estimated such that $p^+ + p^- =1$ $\implies$ $z = MP(\{z^+,z^-\},1)$. $z^+$ and $z^-$ are computed using the reverse water-filling constraints as;
 
\begin{equation}
\sum_{i}^{}[w_{i}^+ + x_{i}^+ - z^+]_+ + [w_{i}^- + x_{i}^- - z^+]_+ + [b^+ - z^+]_+ = \gamma
\label{eq3_per}
\end{equation}
\begin{equation}
\sum_{i}^{}[w_{i}^+ + x_{i}^- - z^-]_+ + [w_{i}^- + x_{i}^+ - z^-]_+ + [b^- - z^-]_+ = \gamma 
\label{eq4_per}
\end{equation}
where  $x_{i}$ is the input sample and $w_{i}$ is the corresponding weight in the log-likelihood domain. 

\subsection{Parameter update rule during training}\label{perc_paraupdate}
Considering a two class problem class$+$ and class$-$, the error function can be written as;
\begin{equation}
E= \sum_{n}^{}|y_{n}^{+}-p^+| + |y_{n}^{-}-p^-|
\label{eq1_per1}
\end{equation}
where 

$y_{n}^{+}$: label for class$+$ for $n^{th}$ sample \\

$y_{n}^{-}$: label for class$-$ for $n^{th}$ sample\\

$y_{n}^{+} + y_{n}^{-} =1$


Using the error gradients obtained from eq. \eqref{eq1_per1}, the weight and bias are updated during each iteration as follows;
\begin{equation}
{}_{}w_{i,(\tau)}^{+}= {}_{}w_{i,(\tau-1)}^{+} - \epsilon \frac{\partial E}{\partial w_{i,(\tau-1)}^+} \\
\label{eq_perc_para1}
\end{equation}
\begin{equation}
{}_{}w_{i,(\tau)}^{-}= {}_{}w_{i,(\tau-1)}^{-} - \epsilon \frac{\partial E}{\partial w_{i,(\tau-1)}^-} \\
\label{eq_perc_para2}
\end{equation}
\begin{equation}
{}_{}b_{(\tau)}^{+}= {}_{}b_{(\tau-1)}^{+} - \epsilon \frac{\partial E}{\partial b_{(\tau-1)}^+} \\
\label{eq_perc_para3}
\end{equation}
\begin{equation}
{}_{}b_{(\tau)}^{-}= {}_{}b_{(\tau-1)}^{-} - \epsilon \frac{\partial E}{\partial b_{(\tau-1)}^-} \\
\label{eq_perc_para4}
\end{equation}
where $\epsilon$ is the learning rate and $\tau$ indicates the iteration step.

Considering eq. \eqref{eq_perc_para1}, it can be proved that,

\begin{equation}
\frac{\partial E}{\partial w_{i}^+} = \sum_{n}^{}sign(p^+ - y_{n}^+)\frac{\partial p^+}{\partial w_{i}^+} + sign(p^- - y_{n}^-)\frac{\partial p^-}{\partial w_{i}^+}
\label{eq5_per}
\end{equation}
where,
\begin{equation}
\frac{\partial p^+}{\partial w_{i}^+}= \left(1-\frac{1}{\textit{A} }\right)\mathbbm{1}(z^+>z) \frac{1}{\textit{A} p}\mathbbm{1}(x_{i}^+ + w_{i}^+ >z^+)
\label{eq8_per}
\end{equation}
Here  $\textit{A} $ indicates the number of $z^+$ such that $z^{+} > z$ and $\mathbbm{1}$ is the indicator function. $\textit{A}p $ indicates the number of elements in the set $\{x_{i}^+ + w_{i}^+ > z^{+}\}$.

The detailed proof and equations for each of the error gradients in  \cref{eq_perc_para1,eq_perc_para2,eq_perc_para3,eq_perc_para4} is given in Appendix \secref{appendixa}.
\subsection{Implementation and results}

\begin{figure*}[htbp]%
	
	
	\begin{subfigure}[t]{0.3\textwidth}
		\includegraphics[width=1.25\linewidth]{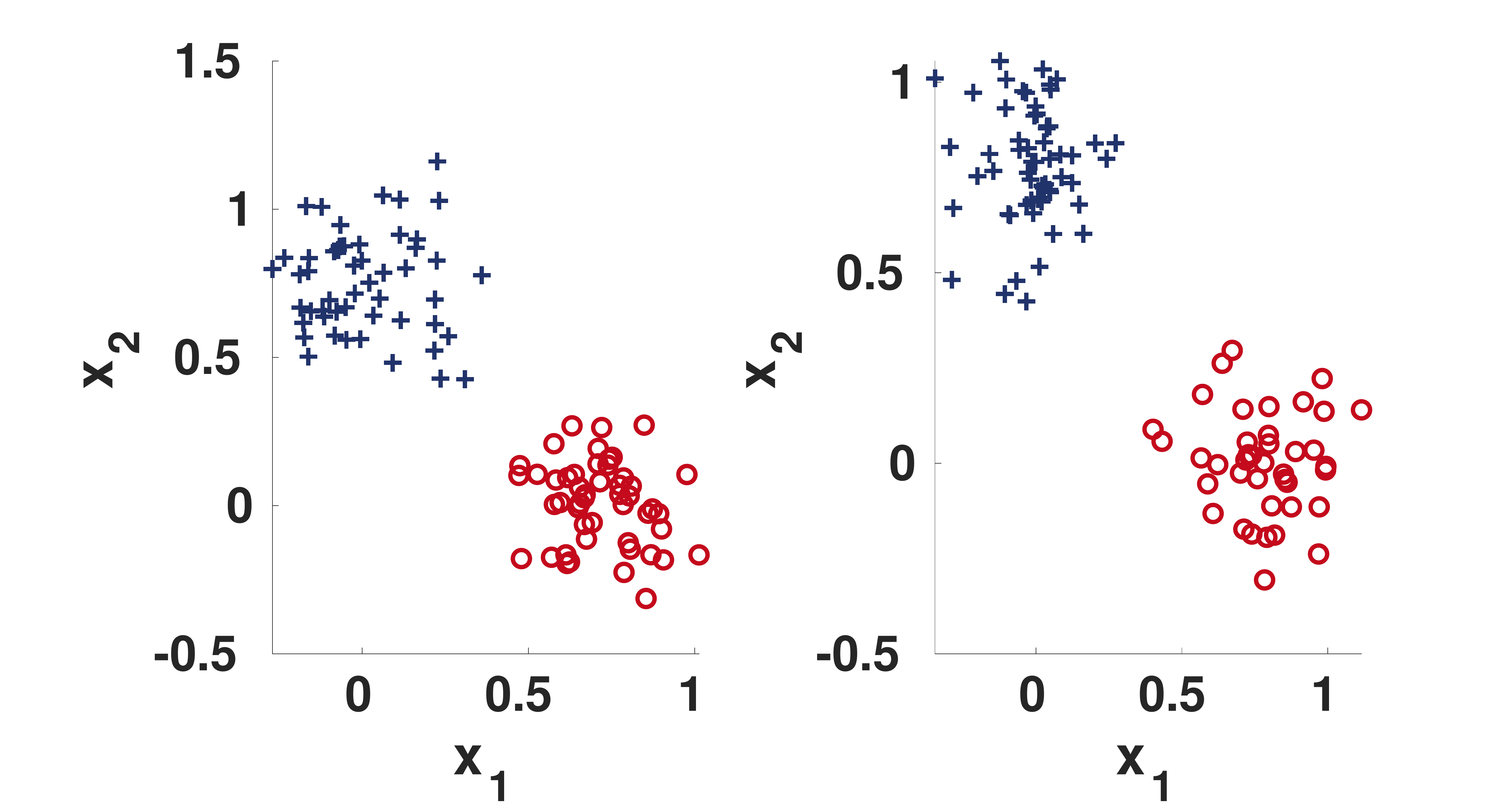}
		\caption{synthetic two class train and test data}
		\label{fig_data_perc}	
	\end{subfigure}\qquad
	\begin{subfigure}[t]{0.3\textwidth}

		\includegraphics[width=\linewidth]{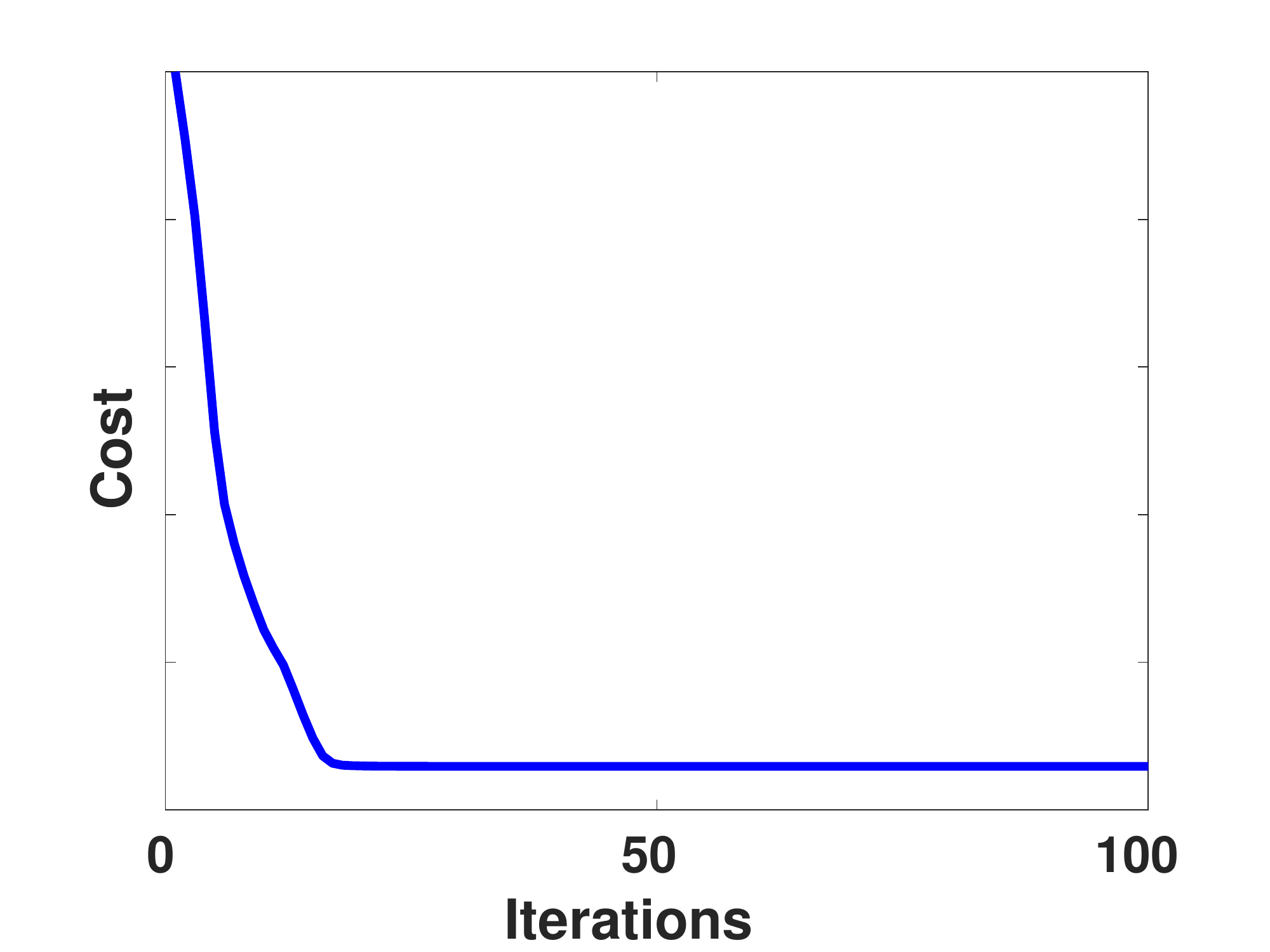}
		\caption{Perceptron training curve}
		\label{fig_train_perc}	
	\end{subfigure}\quad
	\begin{subfigure}[t]{0.3\textwidth}
		\includegraphics[trim=20 0 0 0,width=1.1\linewidth,height=4.2cm]{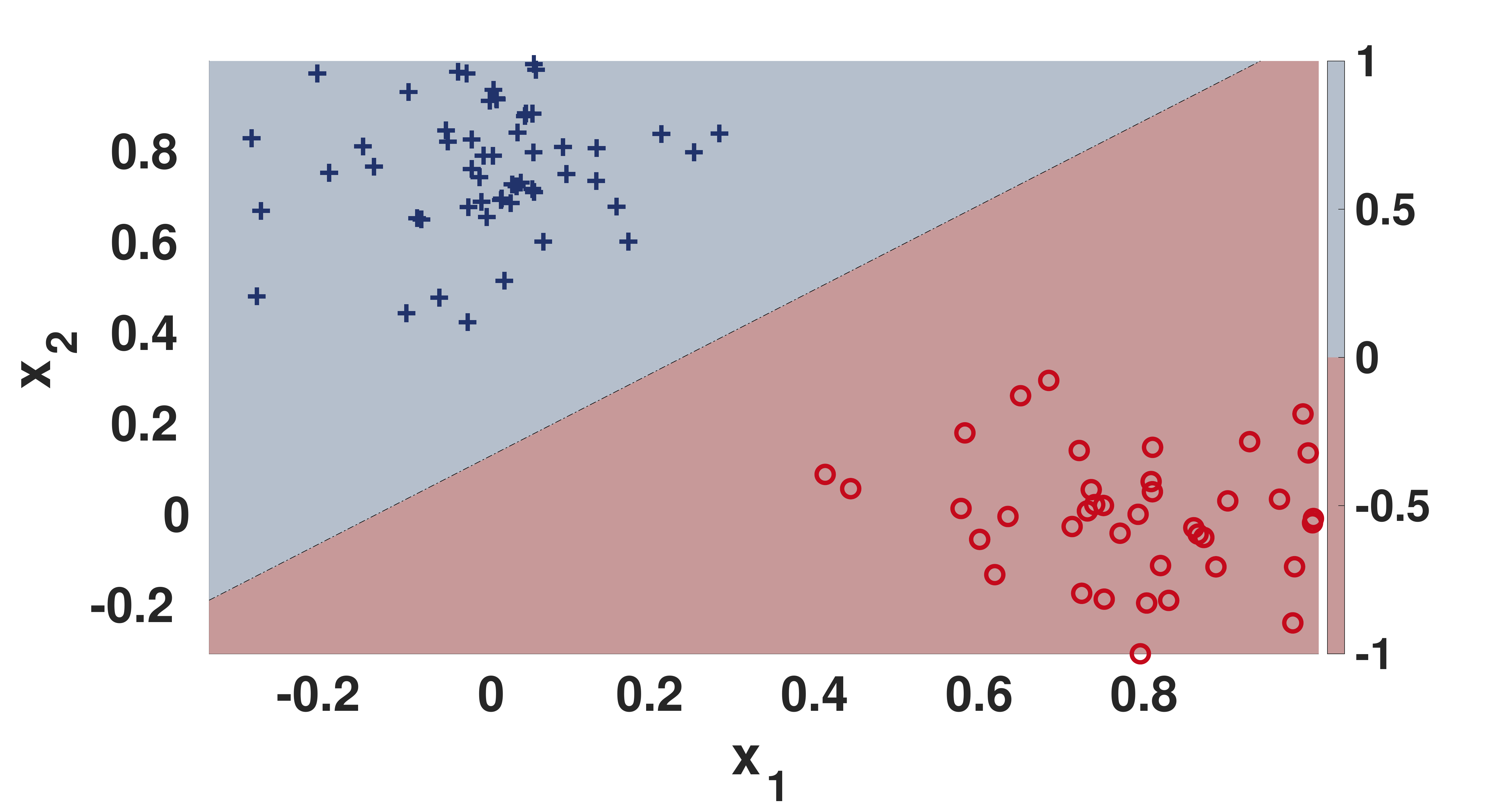}
		\caption{Decision boundary plot of MP based perceptron classification result}
		\label{fig_cont_perc}	
	\end{subfigure}\quad
	
	\caption{}	
\end{figure*}

The formulation is sec. \ref{perceptron} is implemented and results are evaluated using MATLAB. A linearly separable Markovian data is simulated using MATLAB functions for training and testing. We use 100 data samples as train set and 100 samples as test set. 
\subsubsection{Results and discussion}

Figure \ref{fig_data_perc} shows the scatter plot of the linearly separable two class training and test data. The training curve is shown in Fig. \ref{fig_train_perc} which shows that the cost function value reduces  during each iteration. The algorithm gives an accuracy of 100\% as can be seen from the  decision boundary plot of the inference results  in  \cref{fig_cont_perc}.

%
%
%
%
%
%
%
%

\section{Multilayer perceptron based on MP algorithm}\label{mlp}

\begin{figure*}[htbp]%
	
	
	\vspace{-1em}
	\centering

	\includegraphics[height=12cm,width=14cm]{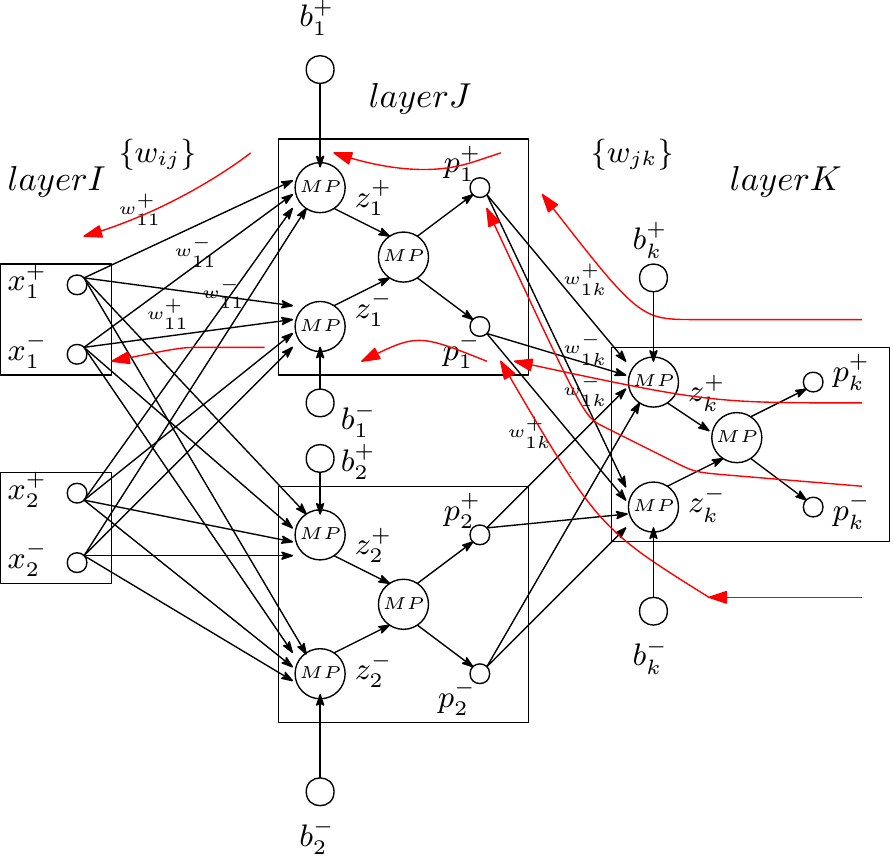}

	
	\caption{A three layer multilayer perceptron (MLP) using MP algorithm as a binary classifier for a non linearly separable xor data; For the present work we use a two dimensional input data and the hidden layer with two nodes.}
	\label{fig_mlp_mp}	
	
\end{figure*}

Figure \ref{fig_mlp_mp} shows an MLP synthesized using MP algorithm.  The network consists of an input layer $I$, a hidden layer $J$ and an output layer $K$ with 2 nodes in the hidden layer. The network parameters are learned by minimizing the $l_1$ norm cost function as shown in \eqref{eq1}. We use an algorithm similar to backpropagation to evaluate the error gradient in-order to update the network parameters. The red arrows indicate the backward propagation
of error information w.r.t the weights $w_{11}^+$ and $w_{11}^-$ .
\subsection{Inference}
Let the input vector in the log-likelihood domain be ${x}= \{x_i\}; 1\leq i\leq N$. Let $\{w_{ij}\}$ and $\{w_{jk}\}$ be the set of learned weights from node $i$ of layer $I$ to the node $j$ in layer $J$ and node  $j$ of layer $J$ to the node in output layer $K$ respectively.

From Fig \ref{fig_mlp_mp} the output in differential form is, 
\begin{equation}
p(x) = p_{k}^+- p_{k}^-
\label{eq33_mlp}
\end{equation}

For the output layer $K$;
\begin{align}
p_{k}^+ & =[z_{k}^+-z_{k}]_+ \nonumber \\
p_{k}^- & =[z_{k}^--z_{k}]_+
\label{eq2}
\end{align}
where $z_k$ is estimated such that $p_{k}^+ + p_{k}^- =1$ $\implies$ $z_{k} = MP(\{z_{k}^+,z_{k}^-\},1)$
and $z_{k}^+$ and $z_{k}^-$  are computed using
\begin{equation}
\sum_{j}^{}[w_{jk}^+ + p_{j}^+ - z_{k}^+]_+ + [w_{jk}^- + p_{j}^- - z_{k}^+]_+ + [b_{k}^+ - z_{k}^+]_+ = \gamma_k 
\label{eq3}
\end{equation}
\begin{equation}
\sum_{j}^{}[w_{jk}^+ + p_{j}^- - z_{k}^-]_+ + [w_{jk}^- + p_{j}^+ - z_{k}^-]_+ + [b_{k}^- - z_{k}^-]_+ = \gamma_k 
\label{eq4}
\end{equation}
Similarly

For the hidden layer $J$;
\begin{align}
p_{j}^+ & =[z_{j}^+-z_{j}]_+ \nonumber \\
p_{j}^- & =[z_{j}^--z_{j}]_+
\label{eq13}
\end{align}
where $z_j$ is estimated such that $p_{j}^+ + p_{j}^- =1$ $\implies$ $z_{j} = MP(\{z_{j}^+,z_{j}^-\},1)$

where, 
\begin{equation}
\sum_{i}^{}[w_{ij}^+ + x_{i}^+ - z_{j}^+]_+ + [w_{ij}^- + x_{i}^- - z_{j}^+]_+ + [b_{j}^+ - z_{j}^+]_+ = \gamma_j 
\label{eq14}
\end{equation}
\begin{equation}
\sum_{i}^{}[w_{ij}^+ + x_{i}^- - z_{j}^-]_+ + [w_{ij}^- + x_{i}^+ - z_{j}^-]_+ + [b_{j}^- - z_{j}^-]_+ = \gamma_j 
\label{eq15}
\end{equation}

\subsection{Parameter update rule during training}
Considering a two class problem class$+$ and class$-$, the error function can be written as;
\begin{equation}
E= \sum_{n}^{}|y_{nk}^{+}-p_{k}^+| + |y_{nk}^{-}-p_{k}^-|
\label{eq1_mlp_1}
\end{equation}
where 

$y_{nk}^{+}$: label for class$+$ for $n^{th}$ sample \\

$y_{nk}^{-}$: label for class$-$ for $n^{th}$ sample\\

$y_{nk}^{+} + y_{nk}^{-} =1$

The weights and bias are updated during each iteration using the error gradients obtained from \eqref{eq1_mlp_1}  as follows;
\begin{equation}
{}_{}w_{ij,(\tau)}^{+}= {}_{}w_{ij,(\tau-1)}^{+} - \epsilon \frac{\partial E}{\partial w_{ij,(\tau-1)}^+} \\
\label{eq_mlp_para1}
\end{equation}
\begin{equation}
{}_{}w_{ij,(\tau)}^{-}= {}_{}w_{ij,(\tau-1)}^{-} - \epsilon \frac{\partial E}{\partial w_{ij,(\tau-1)}^-} \\
\label{eq_mlp_para2}
\end{equation}
\begin{equation}
{}_{}w_{jk,(\tau)}^{+}= {}_{}w_{jk,(\tau-1)}^{+} - \epsilon \frac{\partial E}{\partial w_{jk,(\tau-1)}^+} \\
\label{eq_mlp_para3}
\end{equation}
\begin{equation}
{}_{}w_{jk,(\tau)}^{-}= {}_{}w_{jk,(\tau-1)}^{-} - \epsilon \frac{\partial E}{\partial w_{jk,(\tau-1)}^-} \\
\label{eq_mlp_para4}
\end{equation}
\begin{equation}
{}_{}b_{j,(\tau)}^{+}= {}_{}b_{j,(\tau-1)}^{+} - \epsilon \frac{\partial E}{\partial b_{j,(\tau-1)}^+} \\
\label{eq_mlp_para5}
\end{equation}
\begin{equation}
{}_{}b_{j,(\tau)}^{-}= {}_{}b_{j,(\tau-1)}^{-} - \epsilon \frac{\partial E}{\partial b_{j,(\tau-1)}^-} \\
\label{eq_mlp_para6}
\end{equation}
\begin{equation}
{}_{}b_{k,(\tau)}^{+}= {}_{}b_{k,(\tau-1)}^{+} - \epsilon \frac{\partial E}{\partial b_{k,(\tau-1)}^+} \\
\label{eq_mlp_para7}
\end{equation}
\begin{equation}
{}_{}b_{k,(\tau)}^{-}= {}_{}b_{k,(\tau-1)}^{-} - \epsilon \frac{\partial E}{\partial b_{k,(\tau-1)}^-} \\
\label{eq_mlp_para8}
\end{equation}
where $\epsilon$ is the learning rate and $\tau$ indicates the iteration step.

Considering eq. \eqref{eq_mlp_para1}, it can be shown that,
\begin{equation}
\frac{\partial E}{\partial w_{ij}^+} = \sum_{n}^{}sign(p_{k}^+ - y_{nk}^+)\frac{\partial p_{k}^+}{\partial w_{ij}^+} + sign(p_{k}^- - y_{nk}^-)\frac{\partial p_{k}^-}{\partial w_{ij}^+}
\label{eq16_1}
\end{equation}
where,
\begin{flalign*}
	\frac{\partial p_{k}^+}{\partial w_{ij}^+} & = &  \\
\end{flalign*}
\begin{align}
= \left(1-\frac{1}{\textit{A} _k}\right)\mathbbm{1}(z_{k}^+>z_{k})
\frac{1}{\textit{A} p_k}\mathbbm{1}(p_{j}^+ + w_{jk}^+ >z_{k}^+) \nonumber\\
\left(1-\frac{1}{\textit{A} _j}\right)\mathbbm{1}(z_{j}^+>z_{j})
\frac{1}{\textit{A} p_j}\mathbbm{1}(x_{i}^+ + w_{ij}^+ >z_{j}^+) \nonumber\\
+  \left(1-\frac{1}{\textit{A} _k}\right)\mathbbm{1}(z_{k}^+>z_{k})
\frac{1}{\textit{A} p_k}\mathbbm{1}(p_{j}^- + w_{jk}^- >z_{k}^+) \nonumber\\
\left(1-\frac{1}{\textit{A} _j}\right)\mathbbm{1}(z_{j}^->z_{j})
\frac{1}{\textit{A} n_j}\mathbbm{1}(x_{i}^- + w_{ij}^+ >z_{j}^-)\nonumber\\
\label{eq17_1}
\end{align}

Similarly the rest of the terms can be proven.

The detailed proof for each of the error gradients in \cref{eq_mlp_para1,eq_mlp_para2,eq_mlp_para3,eq_mlp_para4,eq_mlp_para5,eq_mlp_para6,eq_mlp_para7,eq_mlp_para8}
is given in Appendix \secref{appendixb}.

\begin{figure*}[htbp]%
	
	
	\begin{subfigure}[t]{0.3\textwidth}
		\includegraphics[width=1.2\linewidth]{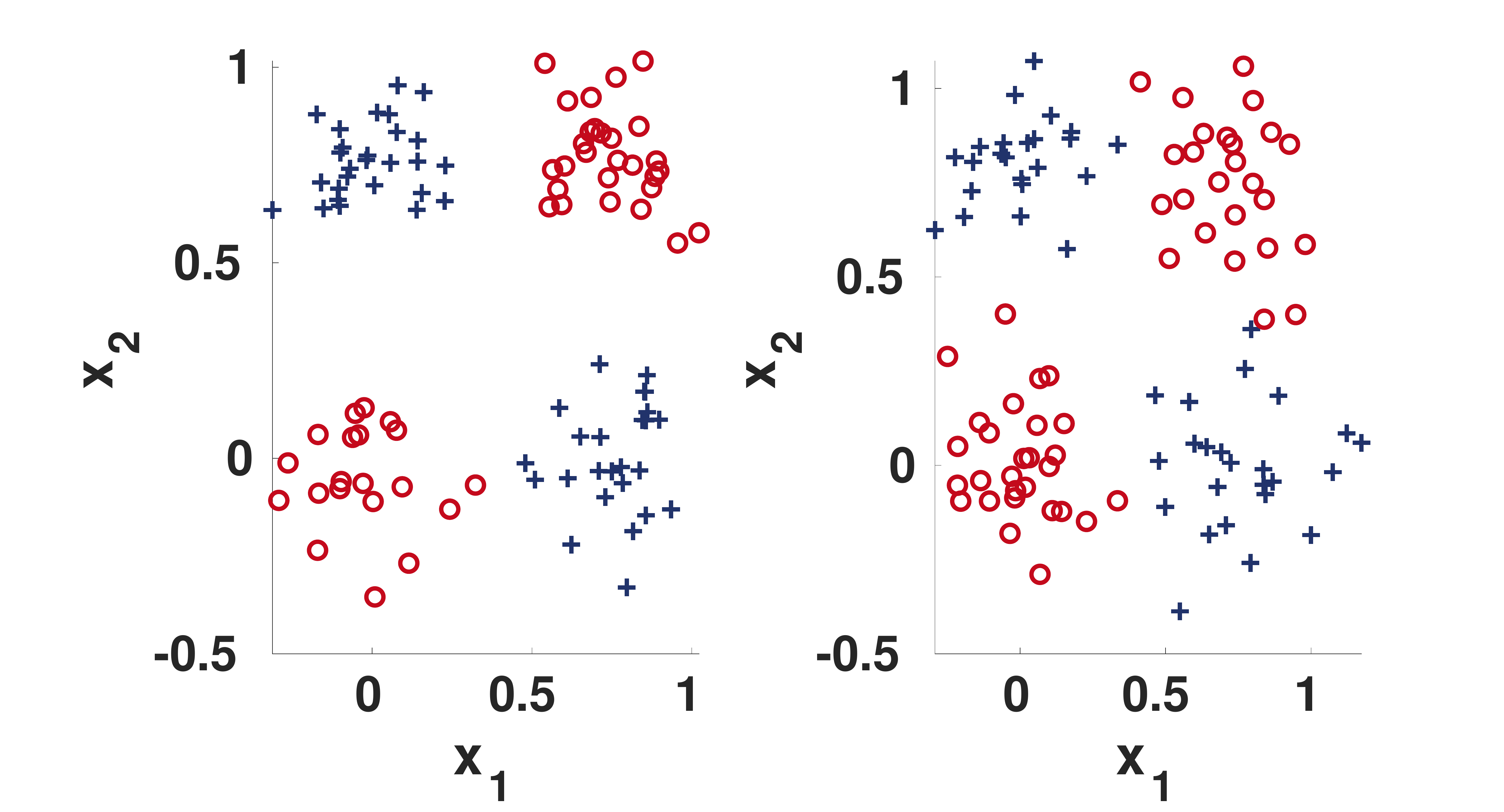}
		\caption{synthetic two class  xor train and test data}
		\label{mlp_data_fig}	
	\end{subfigure}\quad
	\begin{subfigure}[t]{0.3\textwidth}
		
		\includegraphics[width=1\linewidth,height=4cm]{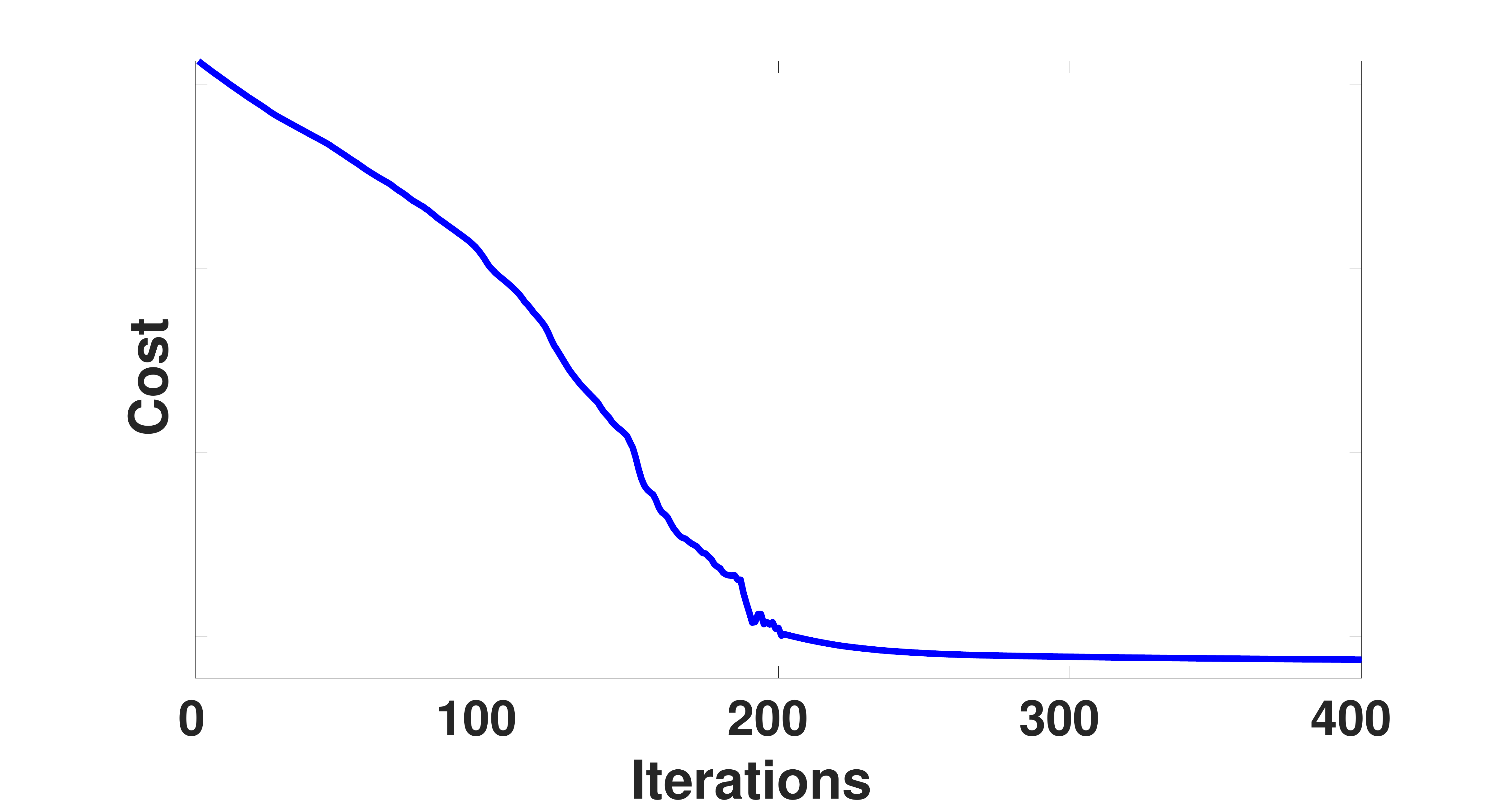}
		\caption{MLP training curve}
		\label{mlp_train_fig}	
	\end{subfigure}\quad
	\begin{subfigure}[t]{0.3\textwidth}
		\includegraphics[width=1\linewidth,height=4cm]{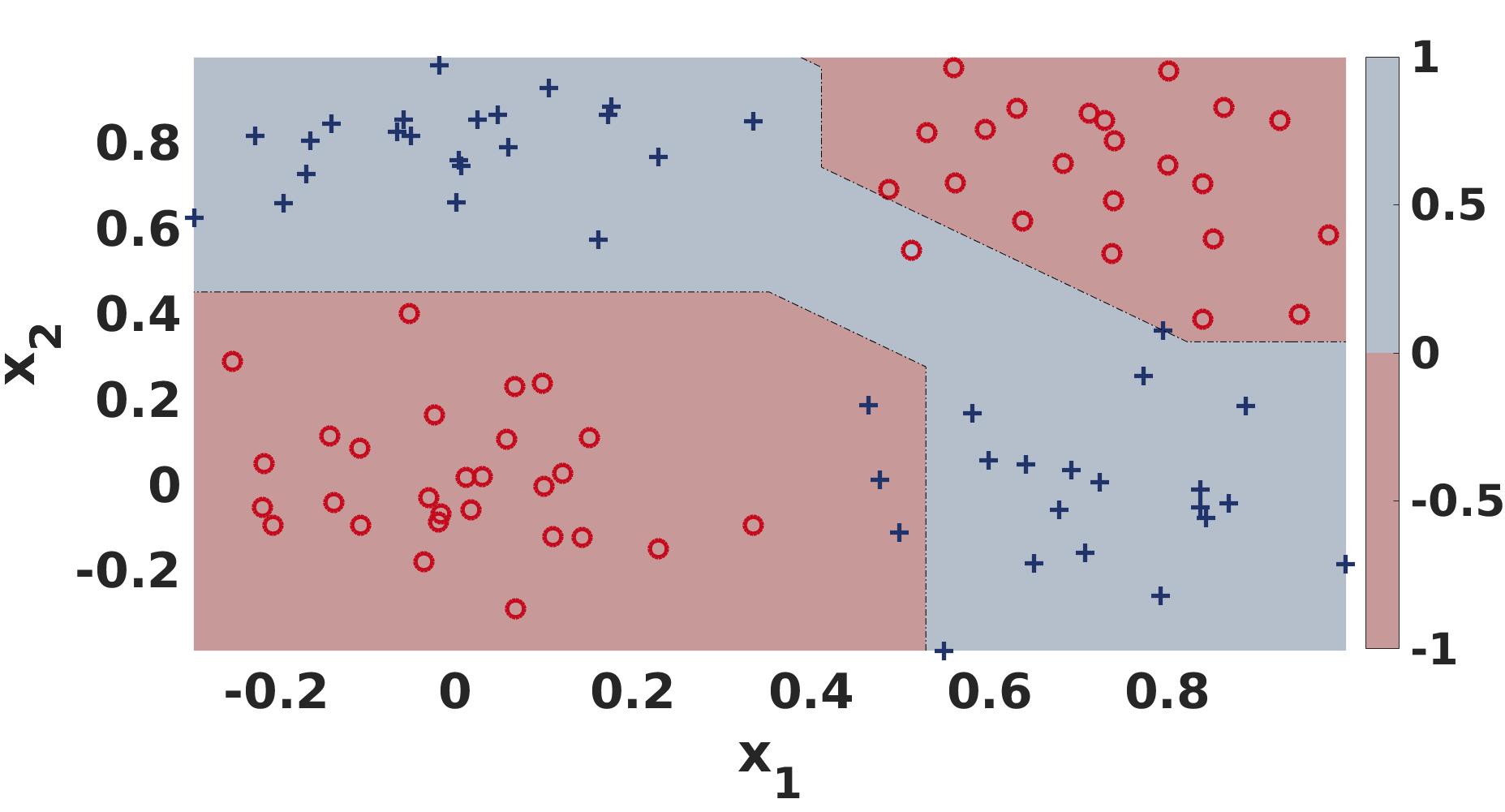}
		\caption{Decision boundary plot of MP MLP classification result}
		\label{mlp_cont_fig}	
	\end{subfigure}%
	%

	\caption{}
	
\end{figure*} 

\subsection{Complexity}
\subsubsection{Training complexity}\label{trcomp}
For a conventional 3 layer MLP with I input nodes, J hidden layer nodes and K output nodes where $K=1$ as shown in \figref{fig_mlp} , the overall training complexity for learning weights is given as,

\begin{flalign}
CT_{MLP-C}= \nonumber\\
& C_A[3JT + 2JIT + JI + J + T] \nonumber\\
&  + C_M[3JT +2JIT  + JT + T] + JT + T
\label{cmlp_final}
\end{flalign}

Similarly it can be proved that for a 3 layer MP MLP with I input nodes, J hidden nodes and $K=1$ output nodes as shown in \figref{fig_mlp_mp}, the overall training complexity for learning weights is,
\begin{flalign}
CT_{MLP-MP}= \nonumber\\
& C_A\big[8JIT + 4JT + 8T + 2J + 2JI\big]   \nonumber\\
& +  C_c \big[2JT \times F \times log(2I)  + 2T \times F \times log(2J)  \nonumber\\
& + 4JIT + 6JT + 2T\big]\nonumber\\
& + C_{S}\big[8JT + 8JIT\big]
\end{flalign}
Here $T$ is the total number of training samples and $F$ is the sparsity factor. $C_A$, $C_c$ and $C_{S}$ are the complexity of addition, comparison and shift operation  respectively. $C_{M}$ indicates the complexity of multiplication  operation.

The detailed proof is given in Appendix \secref{appendixc}.

\begin{figure}[h!]%
	
	
	\begin{subfigure}[t]{0.24\textwidth}
		\includegraphics[width=1.1\linewidth]{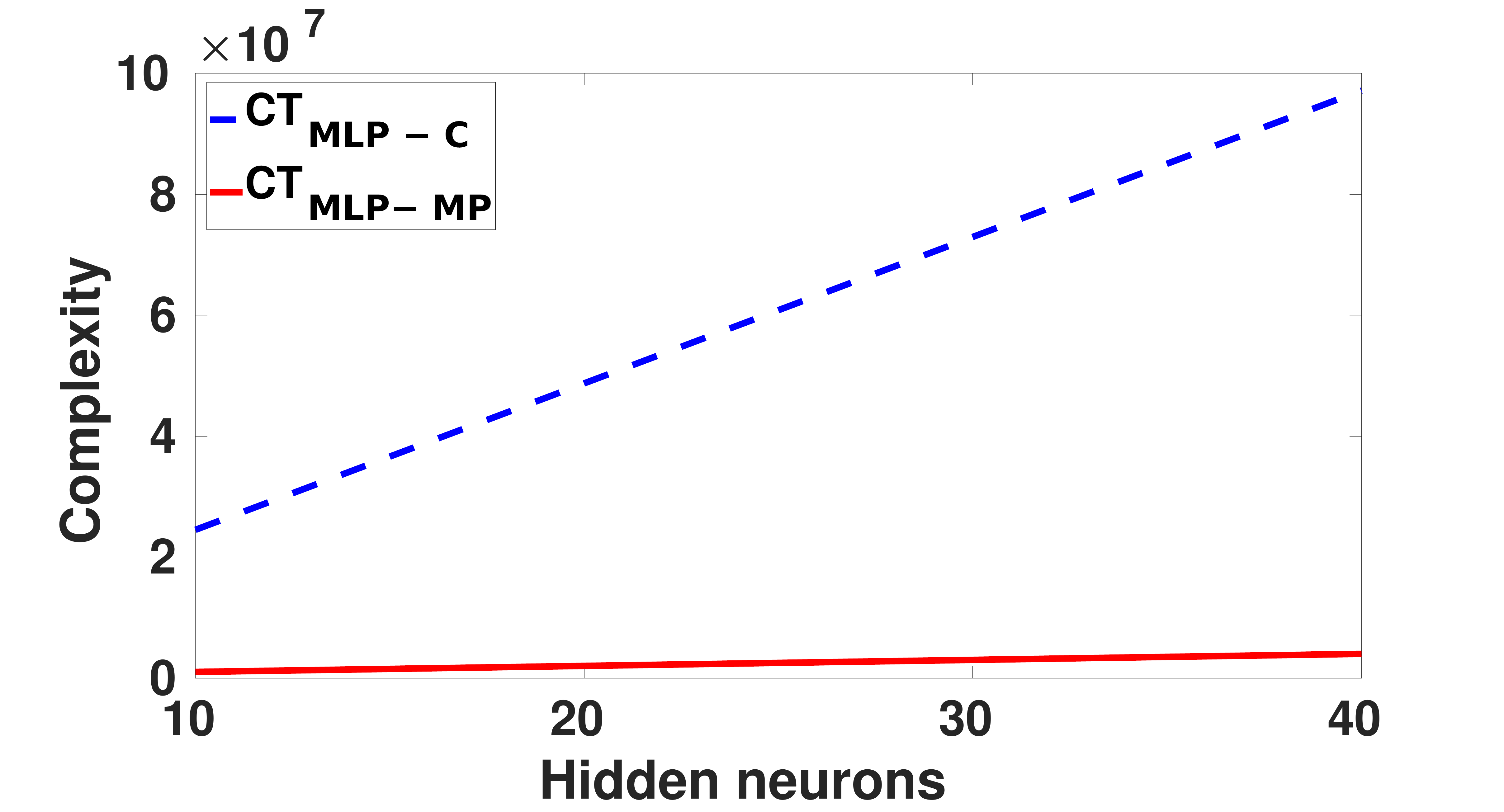}
		\caption{Training complexity, $CT_{MLP-C}$ and $CT_{MLP-MP}$ plots w.r.t hidden neurons}
		\label{comp_mlp1}	
	\end{subfigure}\quad
	\begin{subfigure}[t]{0.24\textwidth}
		
		\includegraphics[width=1.1\linewidth]{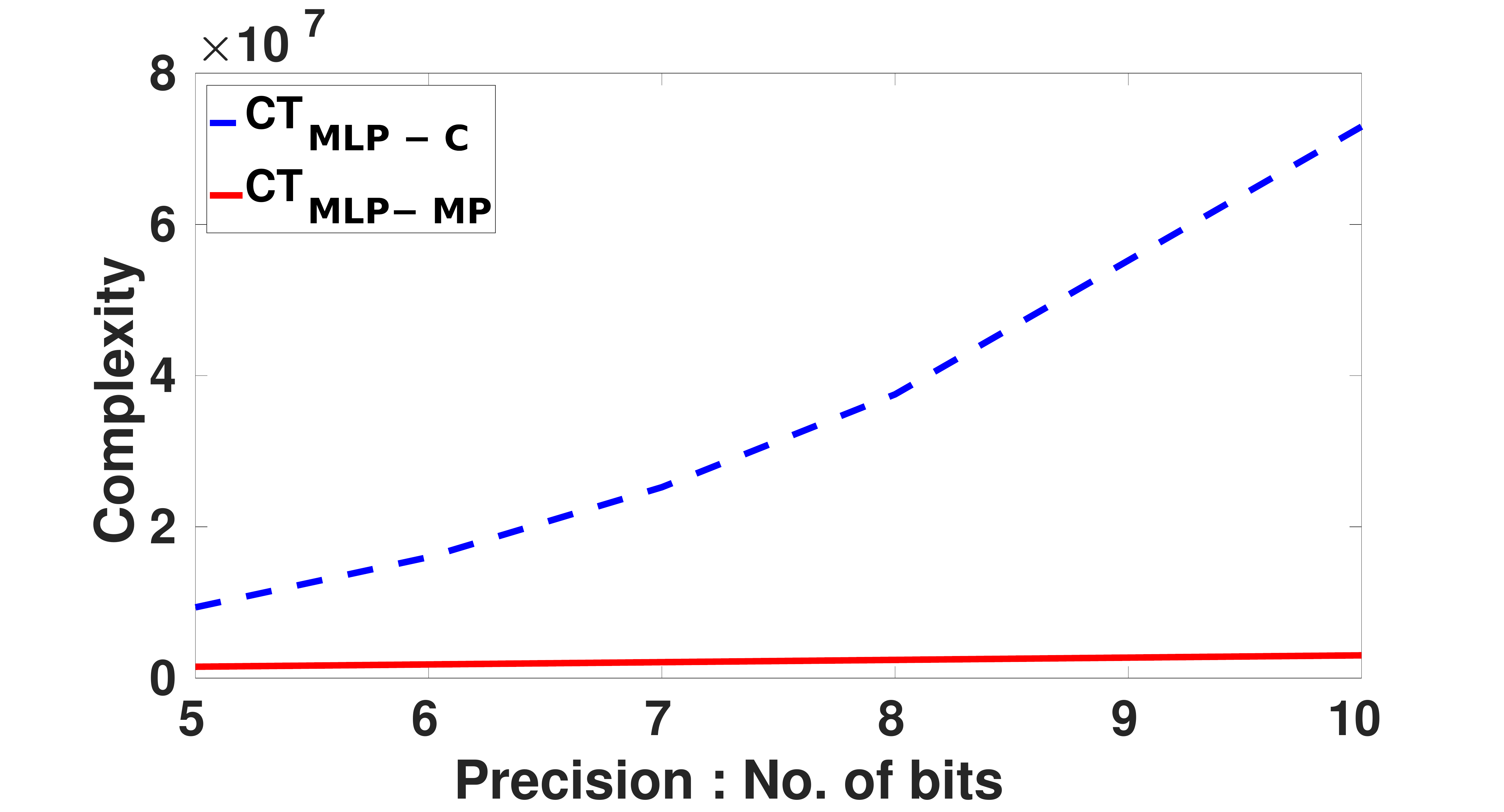}
		\caption{Training complexity, $CT_{MLP-C}$ and $CT_{MLP-MP}$ plots w.r.t precision}
		\label{comp_mlp2}	
	\end{subfigure}\quad
	\begin{subfigure}[t]{0.25\textwidth}
		
		\includegraphics[width=1.2\linewidth]{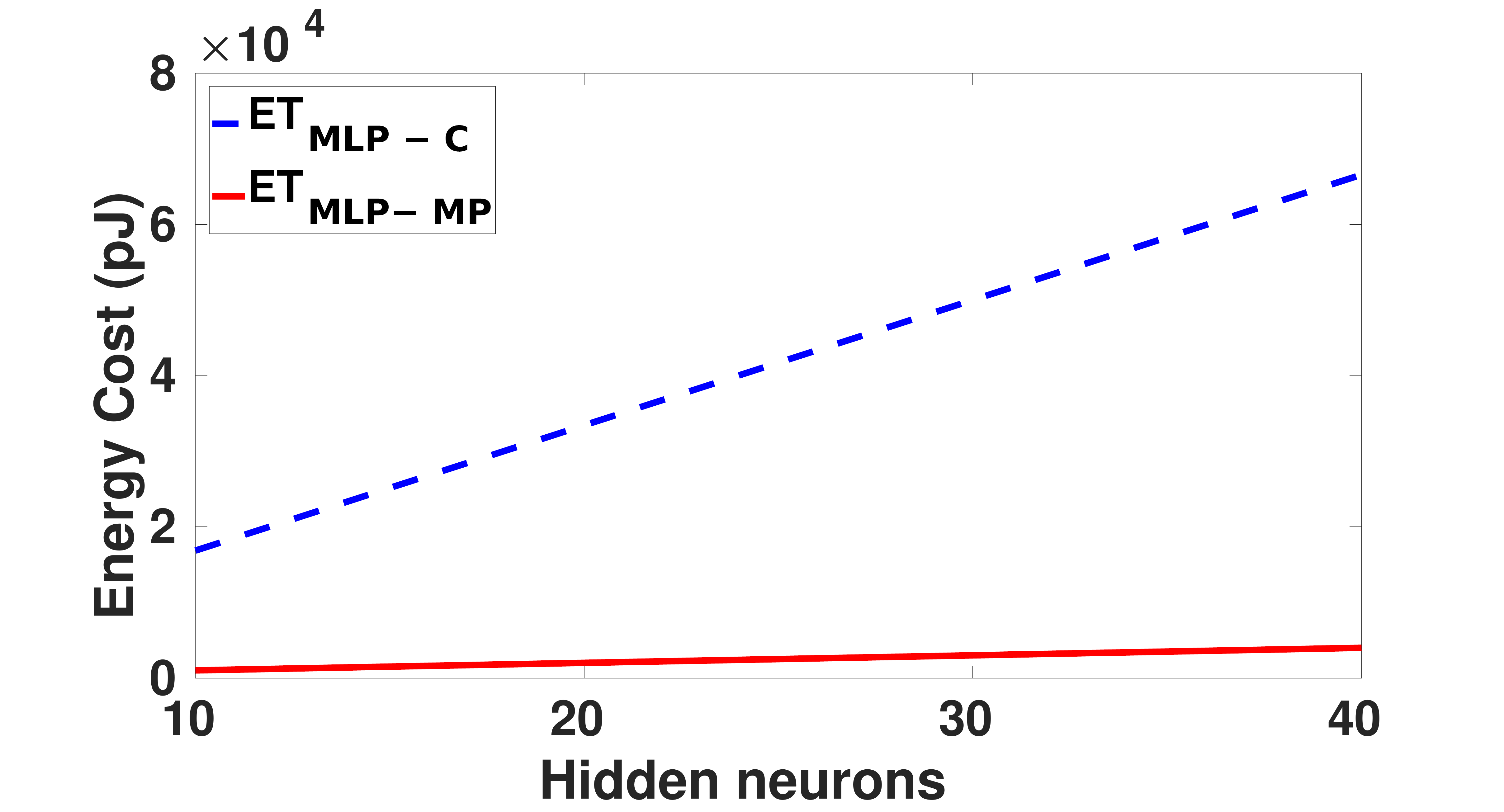}
		\caption{Training energy cost, $ET_{MLP-C}$ and $ET_{MLP-MP}$ plots w.r.t hidden neurons}
		\label{comp_mlp3}	
	\end{subfigure}\quad
	%

	\caption{Variation of training complexity and energy cost}
	
\end{figure}

As per the complexity relation given in \cite{energypaper}, $C_A = 3 \times C_S$ where shift is considered as an elementary operation and 2 complete multiplications require $d^2$ full adders, where $d$ is the number of bits, as explained in \secref{complexity1}.
\Cref{comp_mlp1} shows how the training complexities $CT_{MLP-C}$ and $CT_{MLP-MP}$ varies with the number of hidden neurons $J$. We assume a 10 bit operation for computing the complexities with varying values of $J$. The sparsity factor $F$ is assumed to be 1, which can be further reduced by adjusting the parameter $\gamma$. The input dimension $I$ is assumed to be $2$ similar to the Xor problem discussed in sec. \ref{resultmlpxor} and the number of output neurons $K$ is taken as $1$ considering a two class problem. The number of training samples $T$ is assumed to be $100$. \Cref{comp_mlp2} shows the training complexity variations with varying precisions for the same set of parameters as above and $J=30$. We also show how the energy cost for training varies with the number of hidden neurons for an 8 bit computation as given in \cref{comp_mlp3}. Here we use the energy cost values  in pJ as given in \cite{horowitz20141}.

\subsubsection{Inference complexity}
For a conventional MLP  with the same number of parameters as above, the overall complexity of inference ($CI_{MLP-C}$) for a given  sample $x$ is given as,

\begin{flalign}
CI_{MLP-C}= C_A[JI + J]   + C_M[JI + J] + J
\end{flalign}

Considering \cref{eq3,eq4,eq14,eq15}, the overall complexity of inference for MP-MLP ($CI_{MLP-MP}$) for a given  sample $x$ is given as,
\begin{flalign}
CI_{MLP-MP}= \nonumber\\
& C_A\big[4JI + 4J \big]   \nonumber\\
& +  C_c \big[2J \times F \times log(2I)  + 2 \times F \times log(2J)\big] \nonumber\\
\end{flalign}

\begin{figure}[h!]%
 	
 	
 	\begin{subfigure}[t]{0.24\textwidth}
 		\includegraphics[width=1\linewidth]{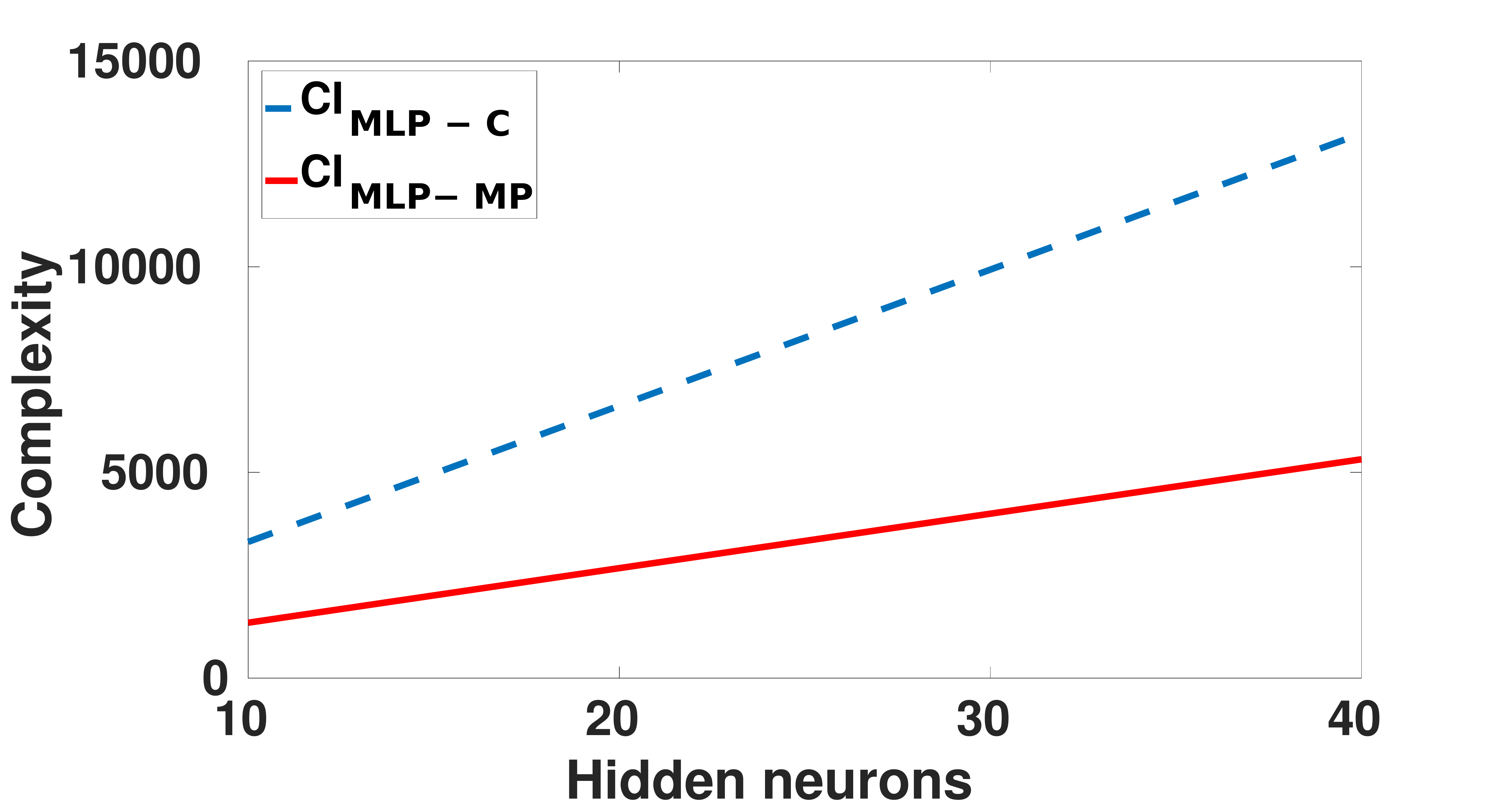}
 		\caption{Inference complexity, $CI_{MLP-C}$ and $CI_{MLP-MP}$ plots w.r.t hidden neurons}
 		\label{comp_mlp_inf}	
 	\end{subfigure}
 	\begin{subfigure}[t]{0.24\textwidth}
 		
 		\includegraphics[width=1\linewidth]{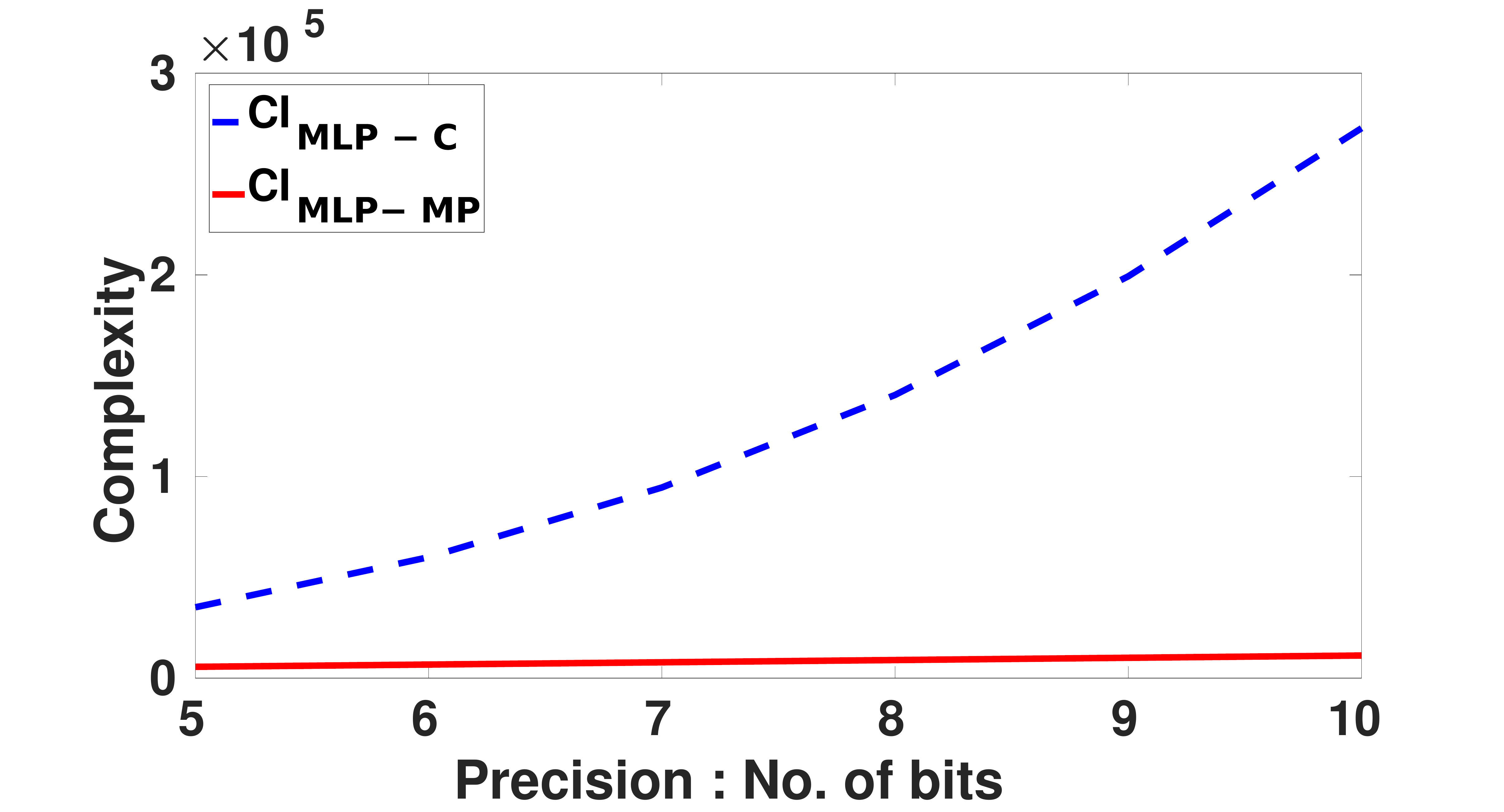}
 		\caption{Inference complexity, $CI_{MLP-C}$ and $CI_{MLP-MP}$ plots w.r.t precision}
 		\label{comp_mlp_inf2}	
 	\end{subfigure}\quad
 	\begin{subfigure}[t]{0.25\textwidth}
 		
 	\includegraphics[width=1\linewidth]{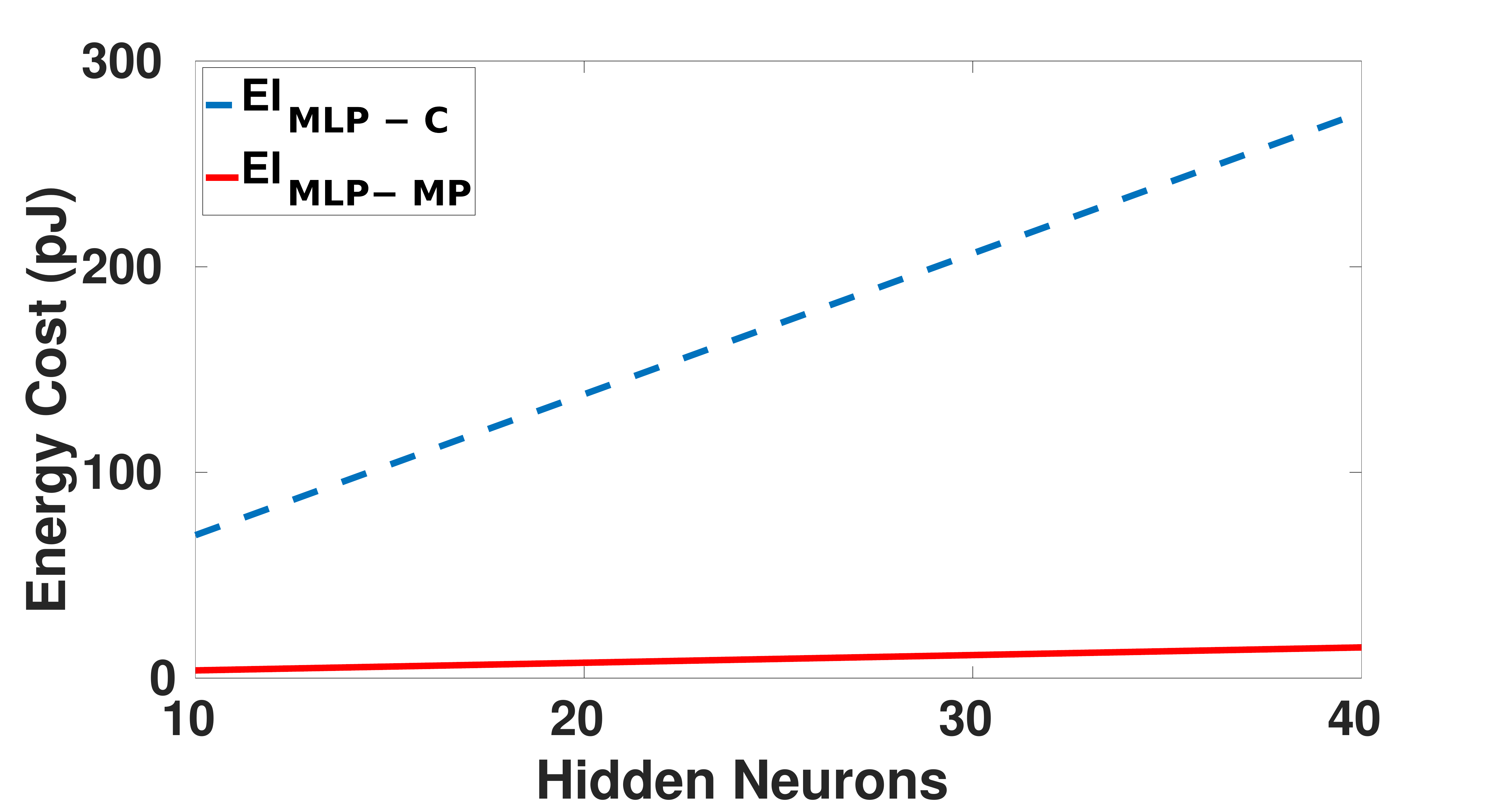}
 	\caption{Inference energy cost, $EI_{MLP-C}$ and $EI_{MLP-MP}$ plots w.r.t hidden neurons}
 	\label{comp_mlp_inf3}	
 	\end{subfigure}\quad
 	%

 	\caption{Variation of inference complexity and energy cost}
 	
 \end{figure}

\figref{comp_mlp_inf} shows how the inference complexities $CI_{MLP-C}$ and $CI_{MLP-MP}$ varies with the number of hidden neurons $J$. Here also we assume a 10 bit operation and consider similar parameters as that of the training case sec. \ref{trcomp}. The inference complexity variation with precision is shown in \cref{comp_mlp_inf2}. \Cref{comp_mlp_inf3} shows the inference energy cost variation with the number of hidden neurons for an 8 bit operation.

It can be inferred from  \cref{comp_mlp1,comp_mlp2} and \cref{comp_mlp_inf,comp_mlp_inf2} that the training and inference complexities of MP-MLP is significantly lower than that of conventional MLP. This will result in  significant improvement of energy cost as shown in \cref{comp_mlp3,comp_mlp_inf3}, as energy per multiplication is more than energy per addition operation \cite{horowitz20141}. As discussed in sec. \ref{complexity1}, this also results in significant reduction in relative area cost. The $L_1$ cost function in conjunction with ReLU operation ensures network sparsity as well.

\subsection{Results on Synthetic Xor data}\label{resultmlpxor}
We use a synthetic non-linearly separable xor data for evaluating our MLP formulation. The train and test set consists of 100 samples each. The network consists of a single hidden layer with 30 neurons and an input and output layer.

The scatter plot of the training and test set is shown in Fig. \ref{mlp_data_fig}. The training curve in Fig. \ref{mlp_train_fig} shows a decreasing cost function per iteration. Figure \ref{mlp_cont_fig} and the table\ref{tab_mlp} shows the classification accuracies of our MLP algorithm on the xor dataset. The algorithm proves to be effective as a non-linear binary classifier as can be seen from the results.
\begin{table}[htbp]
	\resizebox{0.5\textwidth}{!}{
		\begin{tabular}{|c|c|c|c|c|c|c|}
			\hline
			& \multicolumn{3}{c|}{\textbf{Train}}  & \multicolumn{3}{c|}{\textbf{Test}}   \\ \hline
			& Class 1 & Class 2 & \textbf{Overall} & Class 1 & Class 2  & \textbf{Overall}\\ \hline
			Accuracy (\%)        & $100$       & $99$ & $99$             & $99$       & $96$    & $97$   \\ \hline
		\end{tabular}
	}
	\caption{Classification accuracies on the synthetic Xor data using MP MLP}
	\label{tab_mlp}
\end{table}
\subsubsection{Annealing of $\gamma$ parameters}
\begin{figure}[h!]%
	
	
	\begin{subfigure}[t]{0.25\textwidth}
		\includegraphics[width=1\linewidth]{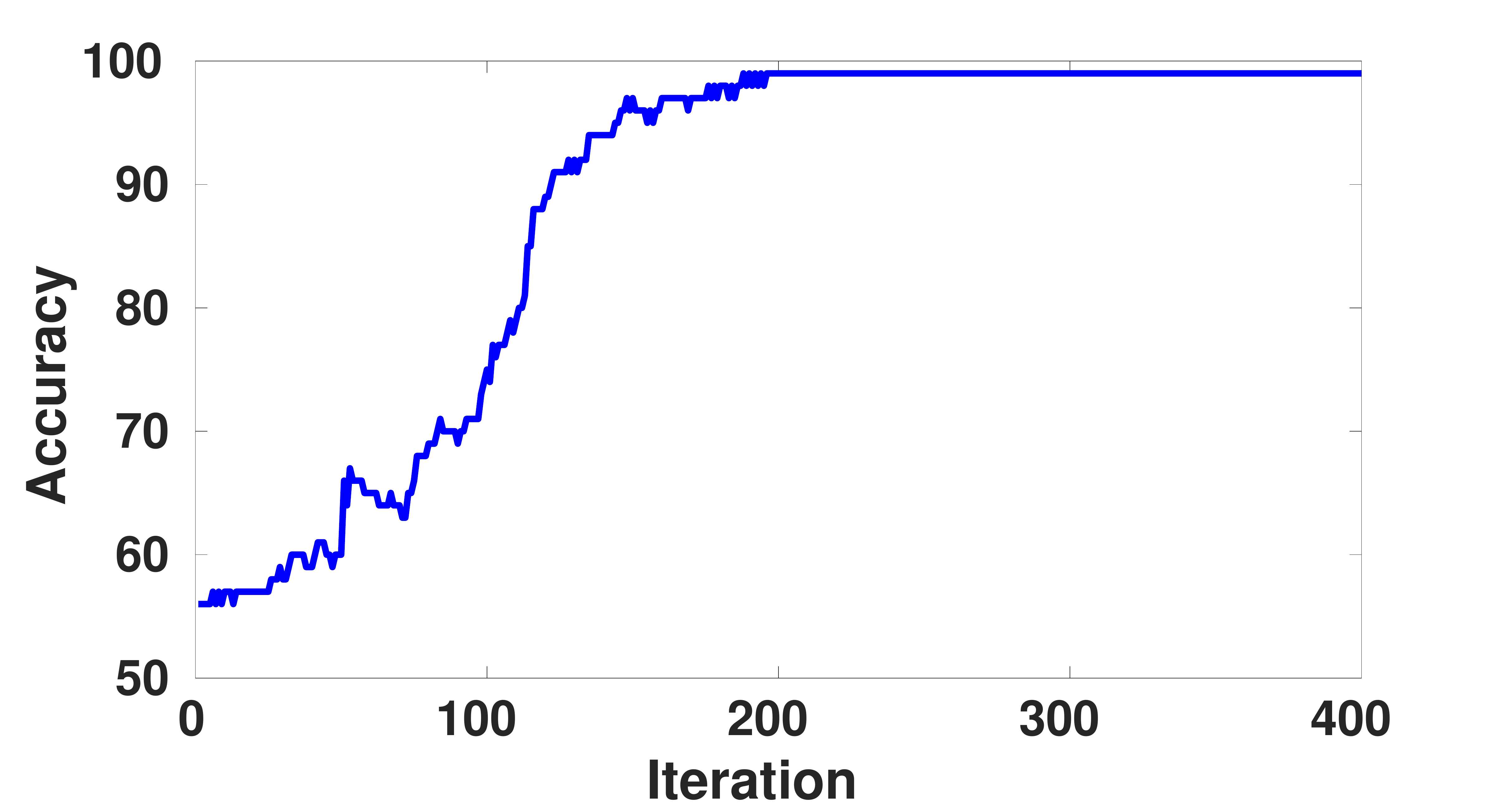}
		\caption{Accuracy variation with $\gamma$ annealing}
		\label{gamma_anneal1}	
	\end{subfigure}\quad
	\begin{subfigure}[t]{0.24\textwidth}
		
		\includegraphics[width=1\linewidth]{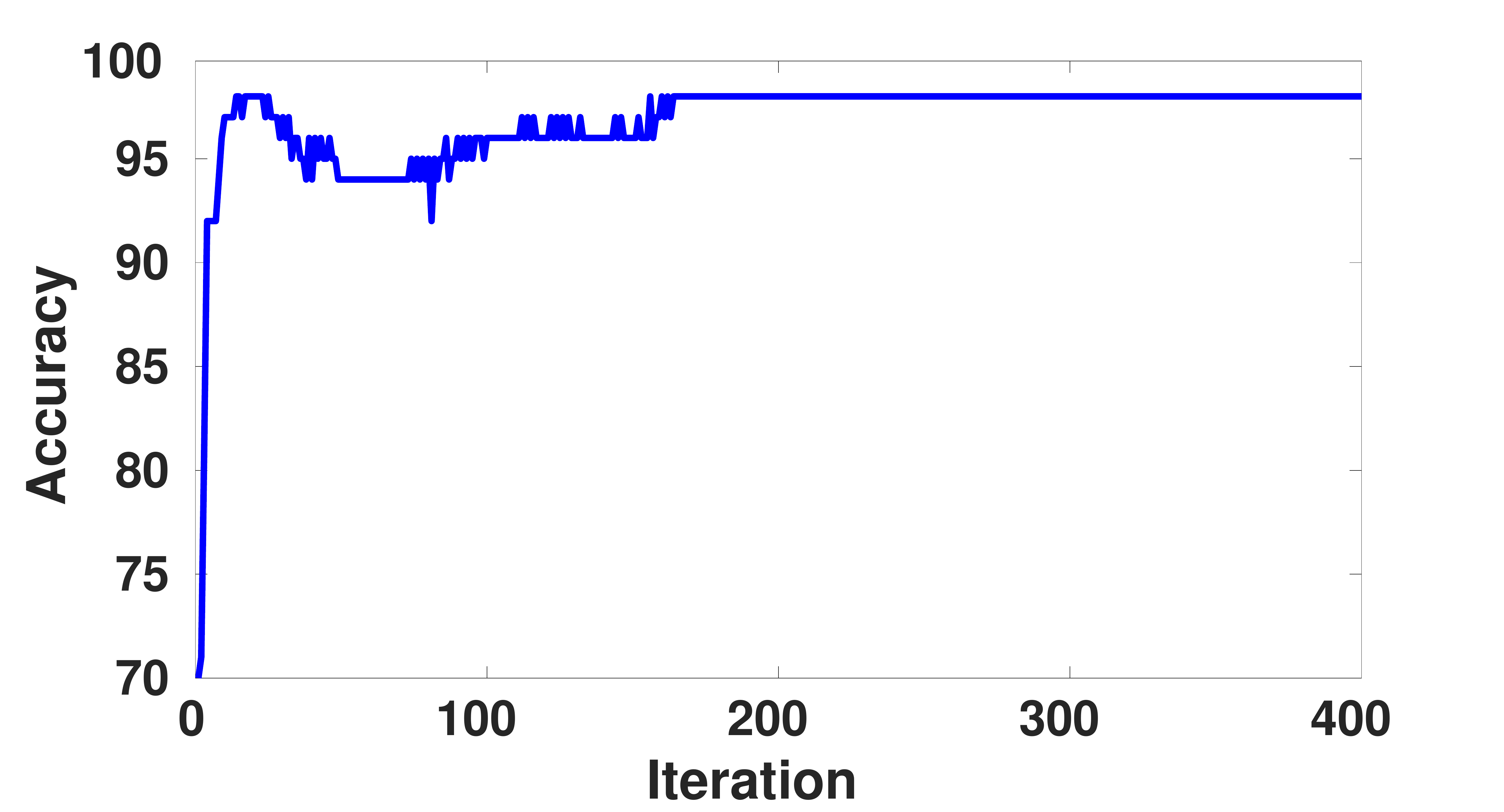}
		\caption{Accuracy variation without $\gamma$ annealing with optimal $\gamma$}
		\label{gamma_anneal2}	
	\end{subfigure}\quad
	\begin{subfigure}[t]{0.24\textwidth}
		
		\includegraphics[width=1\linewidth]{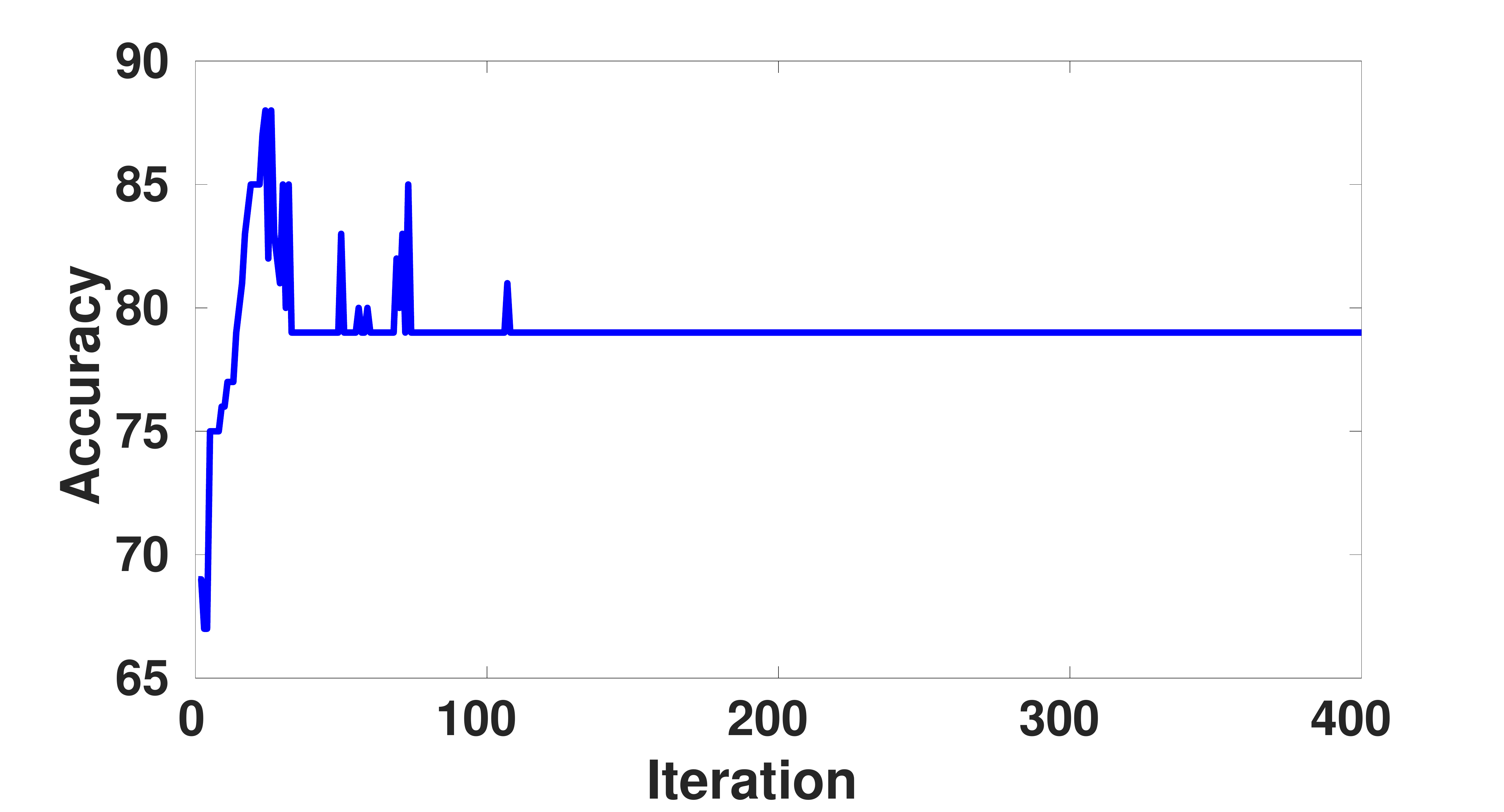}
		\caption{Accuracy variation without $\gamma$ annealing with incorrect $\gamma$}
		\label{gamma_anneal3_wrong}	
	\end{subfigure}
	
	%

	\caption{}
	
\end{figure} 
The values assigned for $\gamma_j$ and $\gamma_k$ play a very crucial role during MP MLP training. The optimal $gamma$ values can be obtained either by a grid search during training and picking the $gamma$ that gives the highest accuracy on the validation data or by updating the $gamma$ parameters with a small step size during each iteration based on the cost function ($\gamma$ annealing). The $gamma$ annealing will also avoid getting stuck at the local minima. The initial value of $\gamma$ and the step size are selected by trial and error depending on the dataset.

 \Cref{gamma_anneal1} shows how the training accuracy varies during each iteration with $\gamma$ annealing for Xor data. \Cref{gamma_anneal2} shows the variation of accuracy when the $\gamma$ values are constant and is an optimal choice. Whereas, \cref{gamma_anneal3_wrong} shows the accuracy variation for a wrong choice of $\gamma$.  It can be seen from  \cref{gamma_anneal1,gamma_anneal2} that the training accuracy increases during each iteration with $\gamma$ annealing and an optimal choice of constant $\gamma$ leading to the highest possible accuracy, whereas a wrong choice of $\gamma$ without any annealing will not give the best possible result.

\subsubsection{Effect of quantization}
\begin{figure}[h!]%
	
	
	\begin{subfigure}[t]{0.5\textwidth}

		\includegraphics[width=1\linewidth]{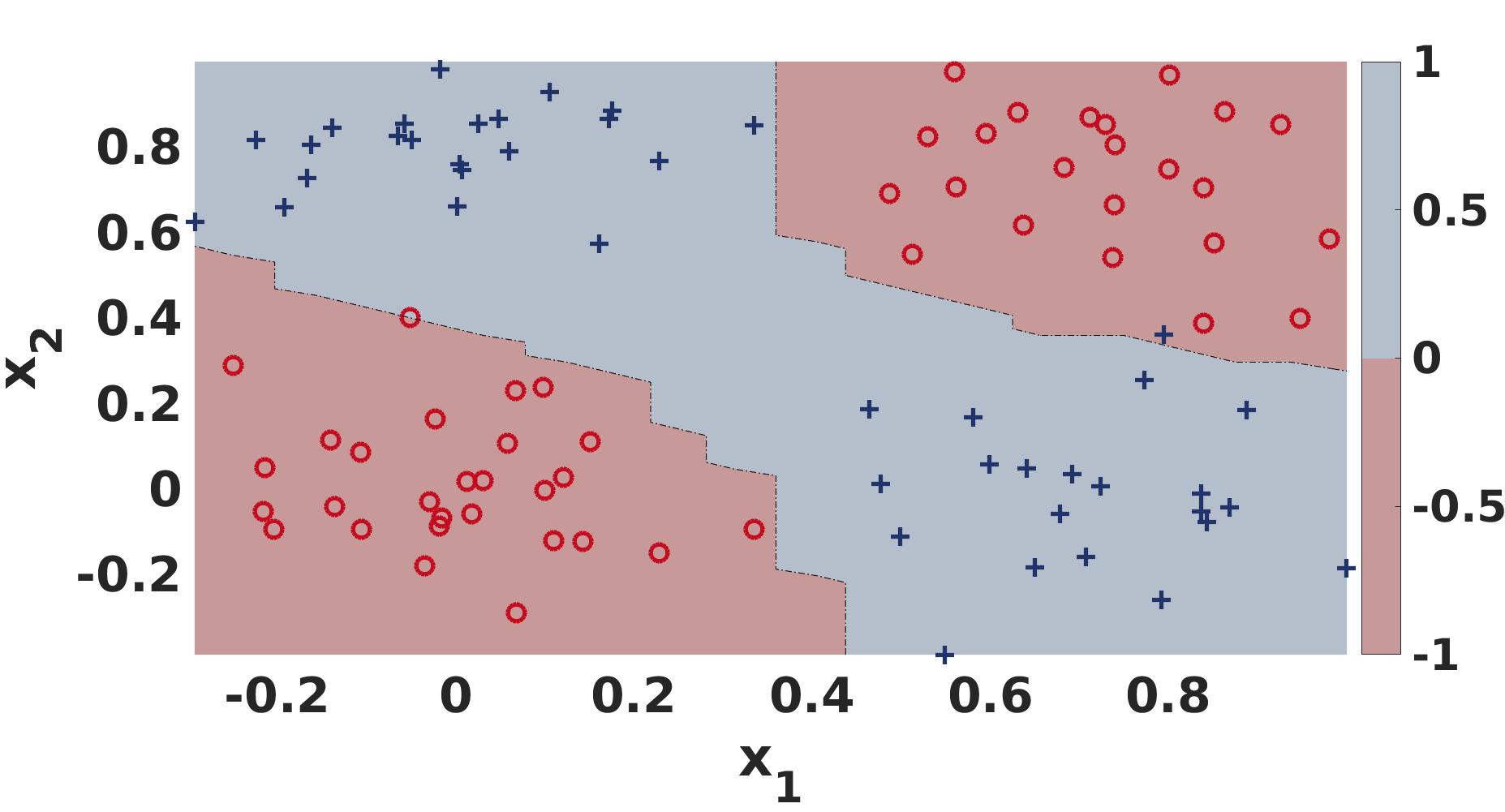}
		\caption{Decision boundary plot of 5-bit fixed point conventional MLP}
		\label{fixed_convmlp}	
	\end{subfigure}
	\begin{subfigure}[t]{0.5\textwidth}
		
		\includegraphics[width=1\linewidth]{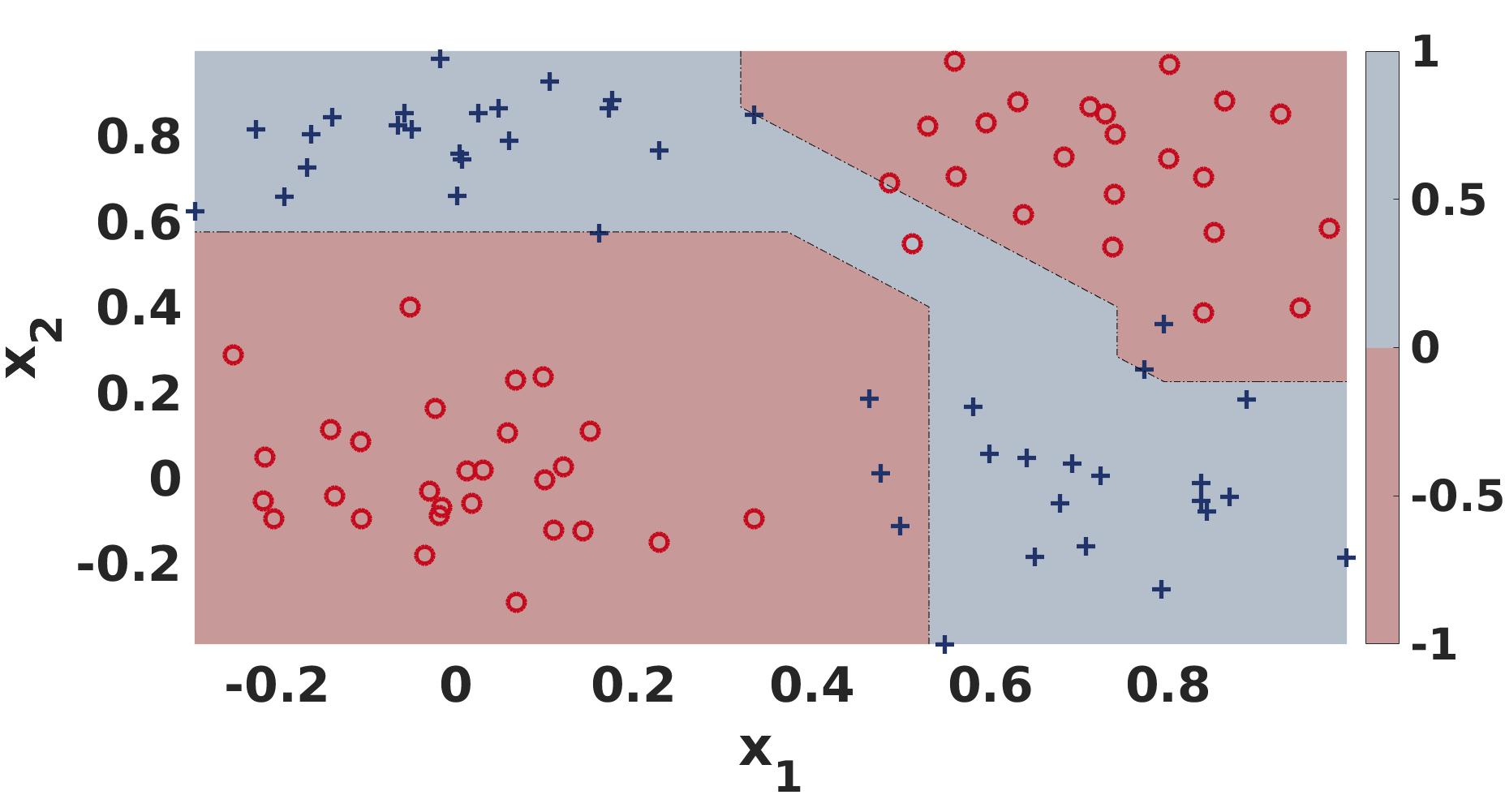}
		\caption{Decision boundary plot of 9-bit fixed point MP MLP}
		\label{fixed_mpmlp}	
	\end{subfigure}
	
	%

	\caption{}
	
\end{figure}

In order to evaluate the effect of quantization, we implemented conventional MLP and MP MLP inference using a fixed point code. The decision boundary of floating point MP MLP shown in Fig. \ref{mlp_cont_fig} has a piecewise linear pattern. \Cref{fixed_convmlp,fixed_mpmlp} show the decision boundary plot using a 5-bit fixed point conventional MLP and 9-bit fixed point MP MLP on Xor data. It can be seen from the figures that fixed point conventional MLP also gives a piecewise linear decision boundary as that of the multiplierless fixed point and floating point MP MLP.

\subsection{ Results on UCI  datasets}
We also evaluate the performances of MP based MLP using single hidden layer on different benchmark UCI  datasets  such as Stalog  Heart,  Diabetes, Wisconsin-breast-cancer and Activity recognition (AReM) datasets \cite{blake1998uci}  and compare the results with  a single hidden layer conventional MLP network with same number of hidden neurons in each case. Table \ref{tab_uci} show the performances in terms of classification accuracy (acc). The dimension (dim) of each dataset is also shown in the table.  From the results shown, it can be inferred that the MP based MLP gives  performances comparable to that of conventional MLP. Thus, MP MLP gives performances at par with that of conventional MLP  with the added benefit of significant reduction in computational complexity and improvement in energy cost as discussed in sec. \ref{complexity1}.

\begin{table*}[htbp]
	\centering
	\resizebox{0.9\textwidth}{!}{  
	\begin{tabular}{|c|c|c|c|c|c|}
		\hline
		& \multicolumn{1}{c|}{Hidden  neurons} & \multicolumn{2}{c|}{\textbf{Conventional MLP}}  & \multicolumn{2}{c|}{\textbf{MP MLP}}   \\ \hline
		&  & Train Acc & Test Acc & Train Acc & Test Acc \\ \hline
		AReM Bending : dim 7  & $15$               & $96\%$   & $95\%$   & $94 \%$       & $95\%$  \\ \hline
		AReM Lying : dim 7 & $15$             & $96 \%$   & $96\% $    & $92 \%$       & $92\%$  \\ \hline
		
		Wisconsin-breast-cancer : dim 10 & $10$             & $96\%$   & $97\%$     & $97\%$       & $98\%$  \\ \hline
		
		Statlog Heart : dim 13 & $25$              & $85 \%$   & $82\% $     & $81 \%$       & $85\%$ \\ \hline
		Pima Indians diabetes : dim 8 & $15$              & $76 \%$   & $73\% $  & $70 \%$       & $73\%$   \\ \hline
	\end{tabular}
		}
	\caption{Classification accuracies on UCI dataset using MP based MLP}
	\label{tab_uci}
\end{table*}
\subsubsection{Effect of quantization}
\begin{figure}[h!]%
	
	
	\begin{subfigure}[t]{0.22\textwidth}
		\includegraphics[width=1.2\linewidth]{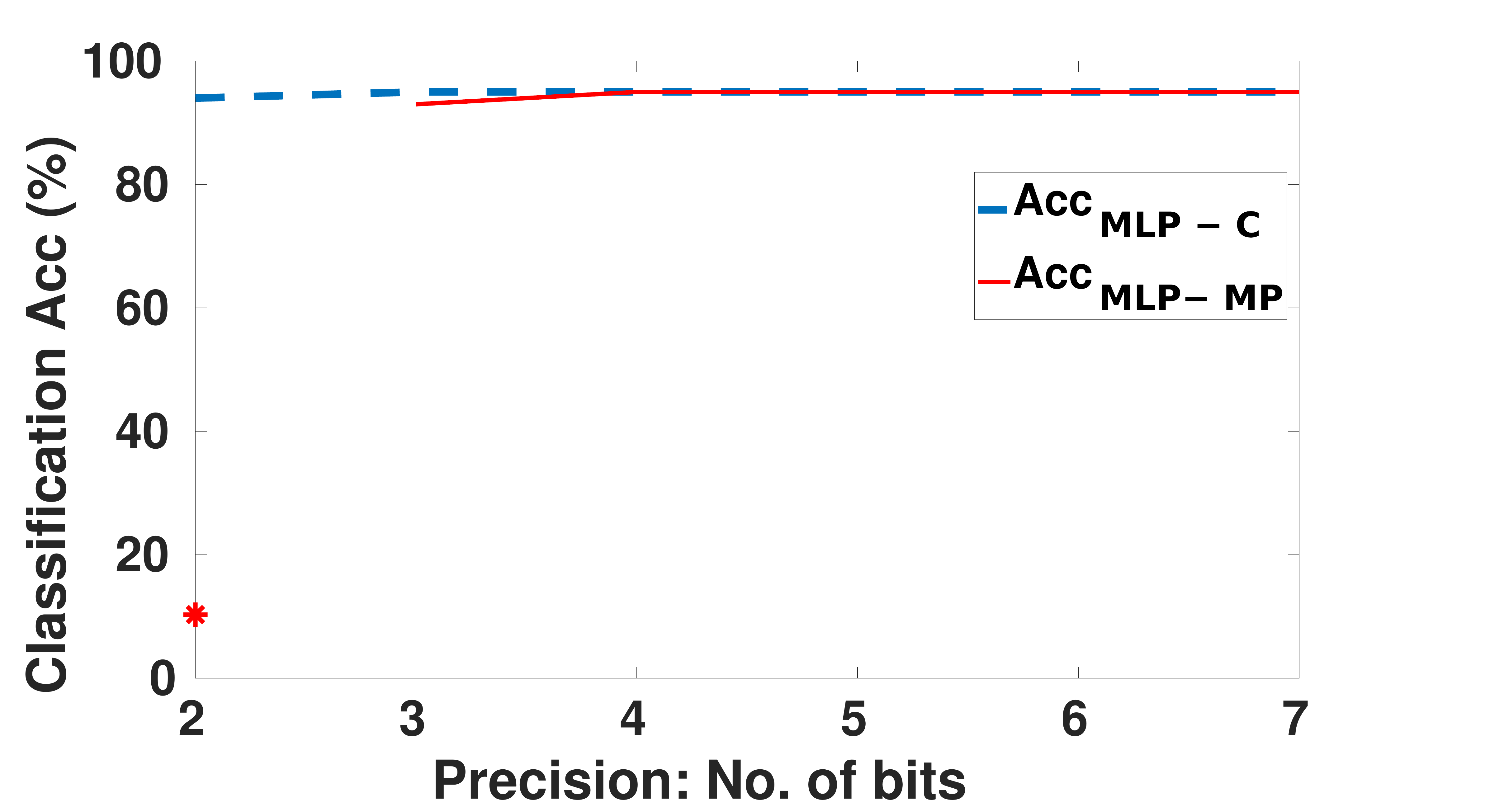}
		\caption{AReM Bending}
		\label{accvsbits1}	
	\end{subfigure}\qquad
	\begin{subfigure}[t]{0.22\textwidth}
		
		\includegraphics[width=1.2\linewidth]{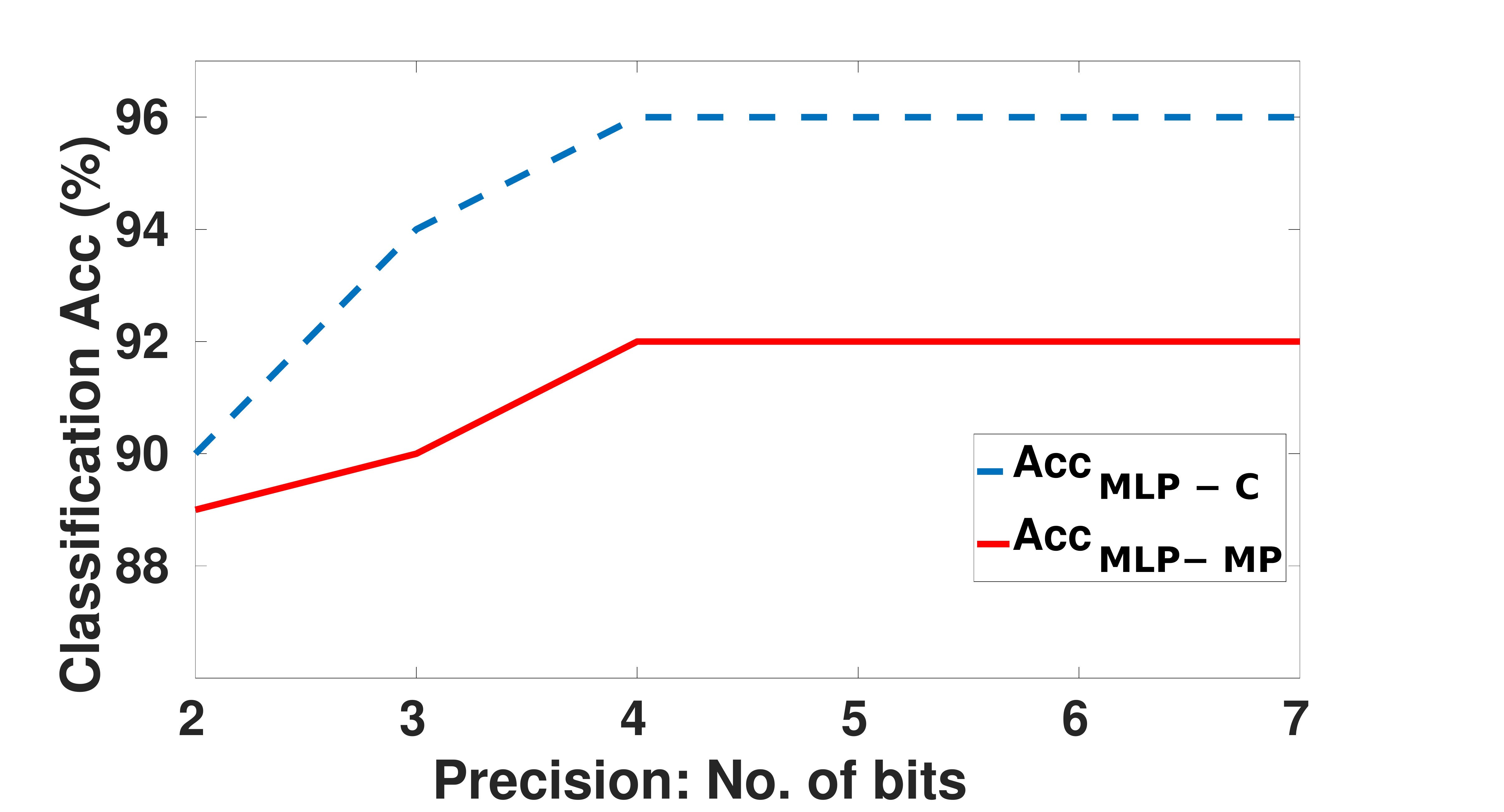}
		\caption{AReM Lying}
		\label{accvsbits2}	
	\end{subfigure}\qquad
	\begin{subfigure}[t]{0.22\textwidth}
		\centering
		\includegraphics[width=1.2\linewidth]{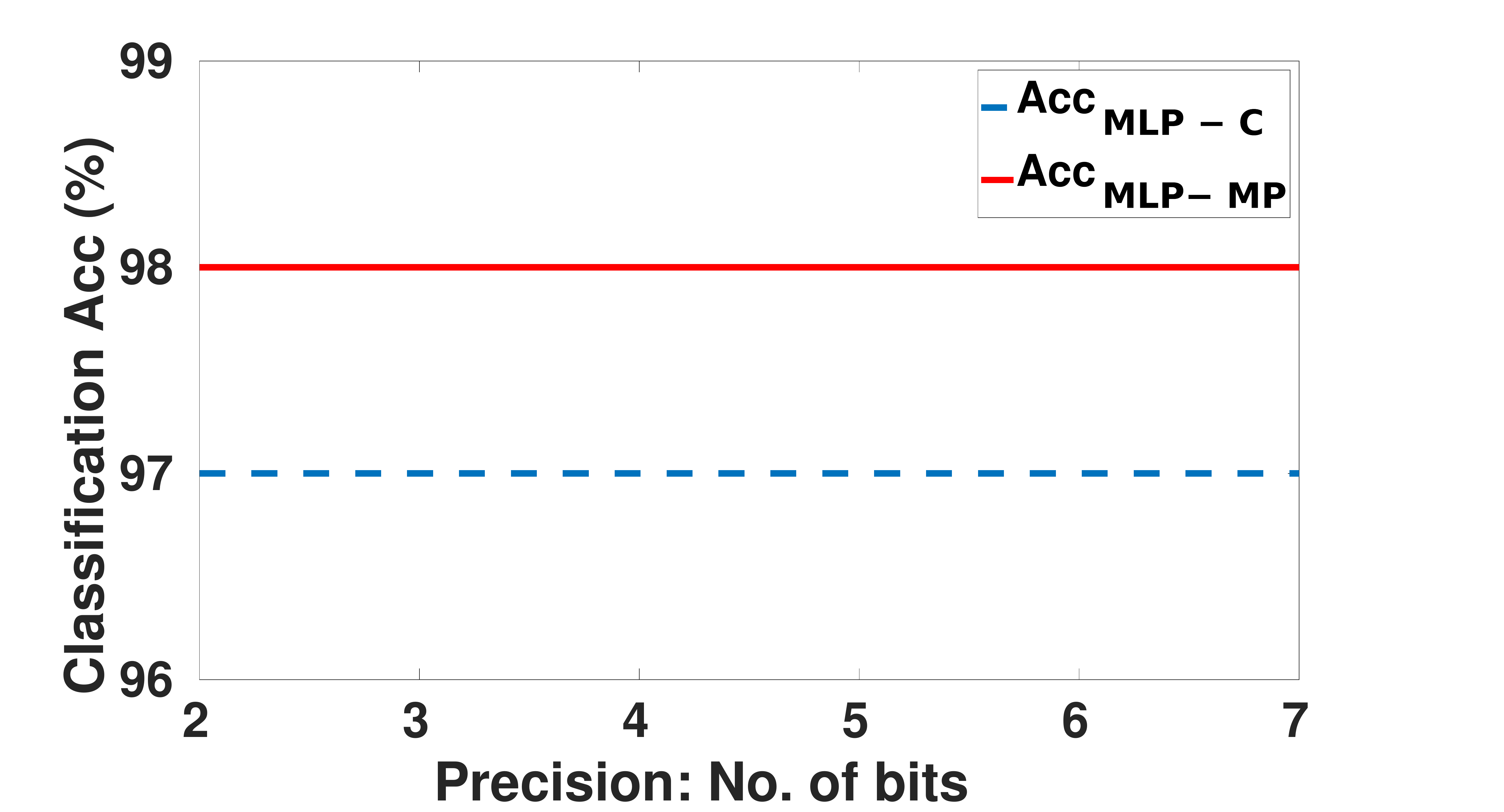}
		\caption{Wisconsin-breast-cancer}
		\label{accvsbits3}
	\end{subfigure}\qquad
	\begin{subfigure}[t]{0.22\textwidth}
		\centering
		\includegraphics[width=1.2\linewidth]{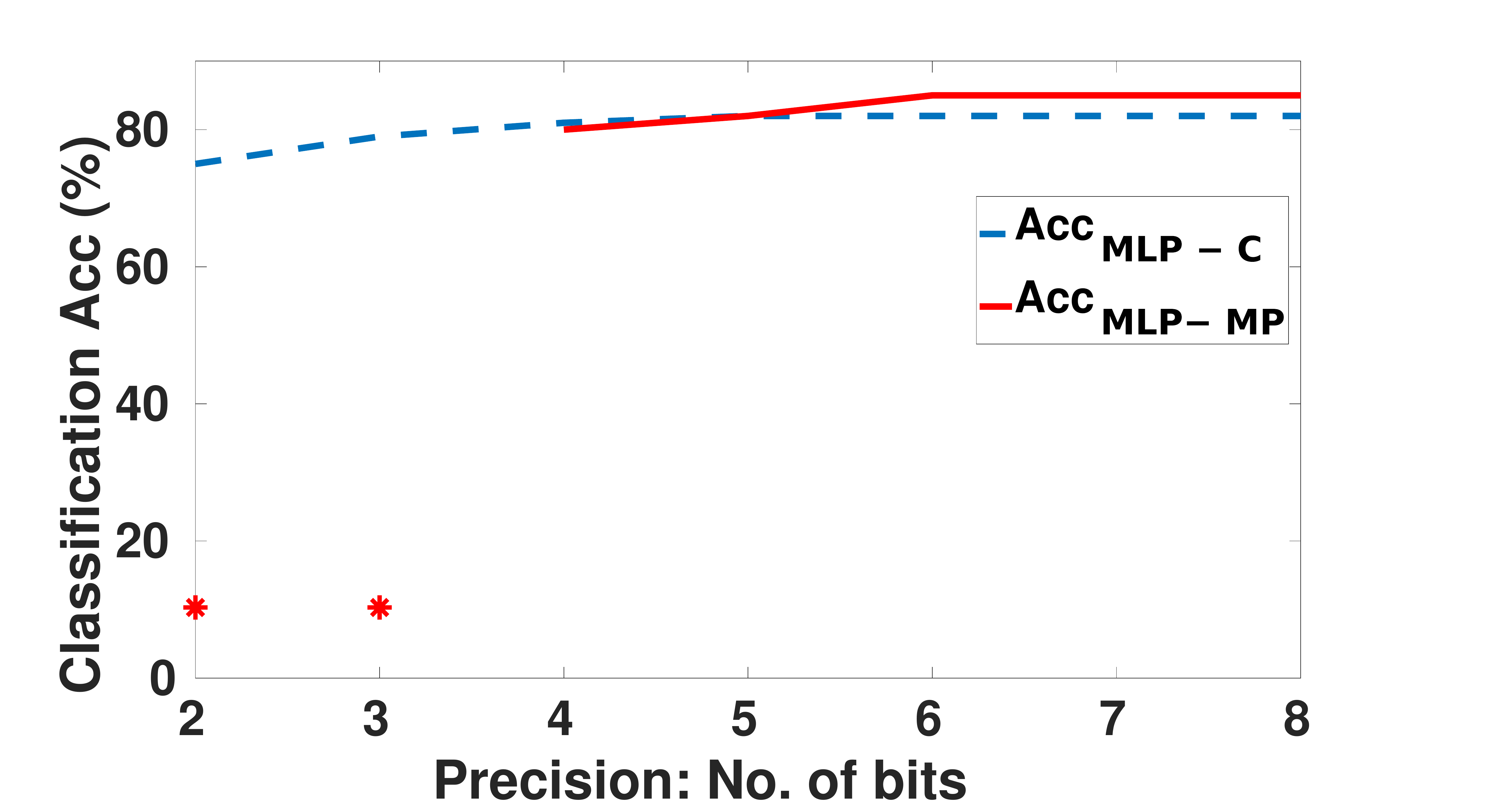}
		\caption{Statlog Heart}
		\label{accvsbits4}
	\end{subfigure}\qquad
	\begin{subfigure}[t]{0.22\textwidth}
		\centering
		\includegraphics[width=1.2\linewidth]{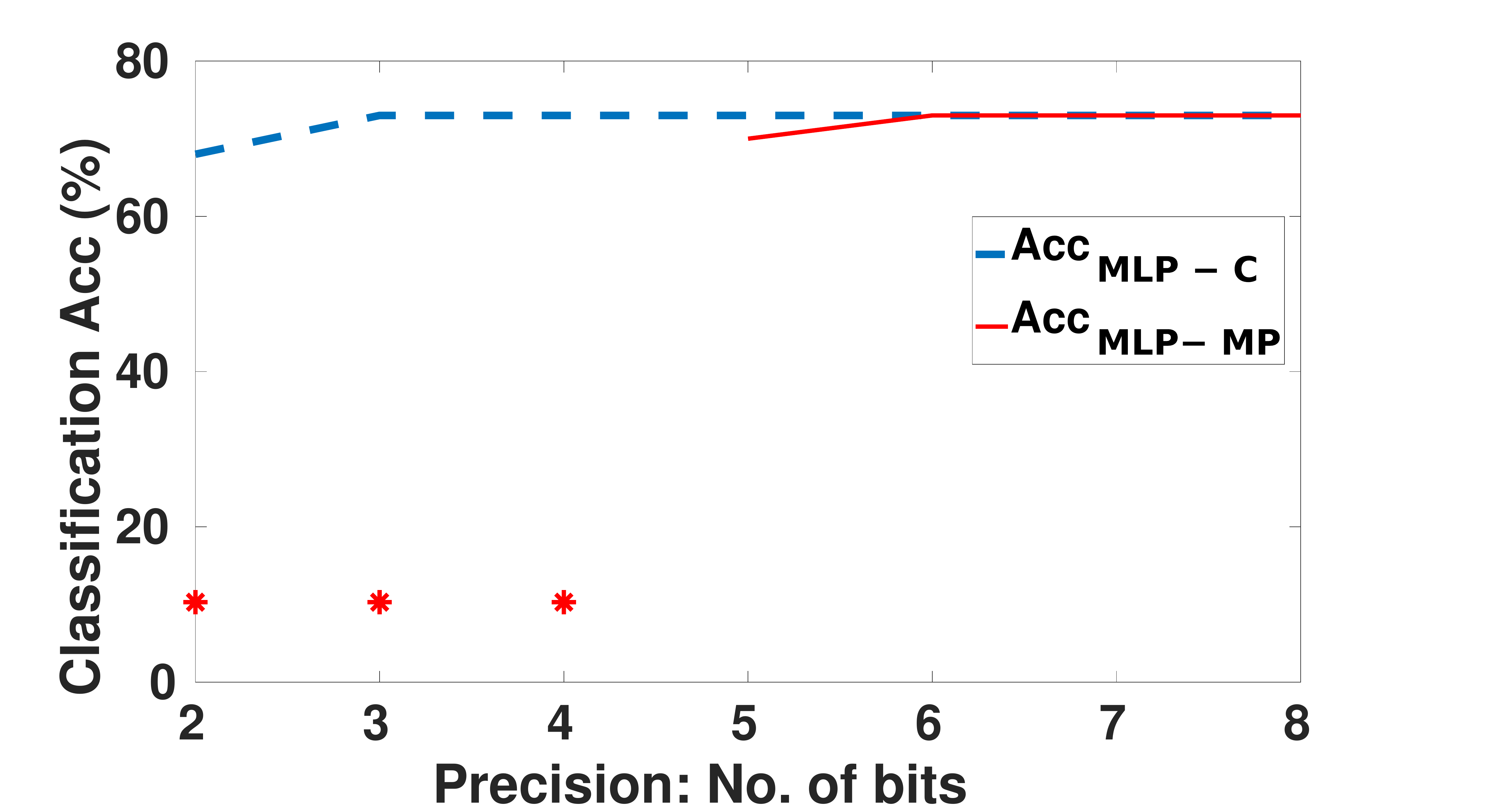}
		\caption{Pima Indians diabetes}
		\label{accvsbits5}
	\end{subfigure}%
	%

	\caption{Variation of $Acc_{MLP-C}$ and $Acc_{MLP-MP}$ with varying precisions at the inference stage. *'s indicate the cases where the results are really poor or inconclusive after multiple runs.}
	
\end{figure}

In order to evaluate the effect of quantization on classification accuracy, we plot the variation of classification accuracy with varying precisions at the inference stage. \Cref{accvsbits1,accvsbits2,accvsbits3,accvsbits4,accvsbits5} show the variation of accuracy values for conventional MLP ($Acc_{MLP-C}$) and MP MLP ($Acc_{MLP-MP}$) with varying precisions at the inference stage (2-bit fixed point to 9-bit fixed point) for UCI datasets. It can be seen that for some cases, MP MLP gives inconclusive or poor accuracies at lower precision of 2-bit and 3-bit fixed point computations. This could be due to the effect of approximation errors in MP formulation which needs further analysis. However, at higher precisions above 4-bit, MP MLP gives performances at par with that of conventional MLP.

\section{SVM based on MP algorithm}\label{svm}
\begin{figure*}[h!]%
	
	
	\begin{subfigure}[t]{0.3\textwidth}
		\includegraphics[width=1.1\linewidth]{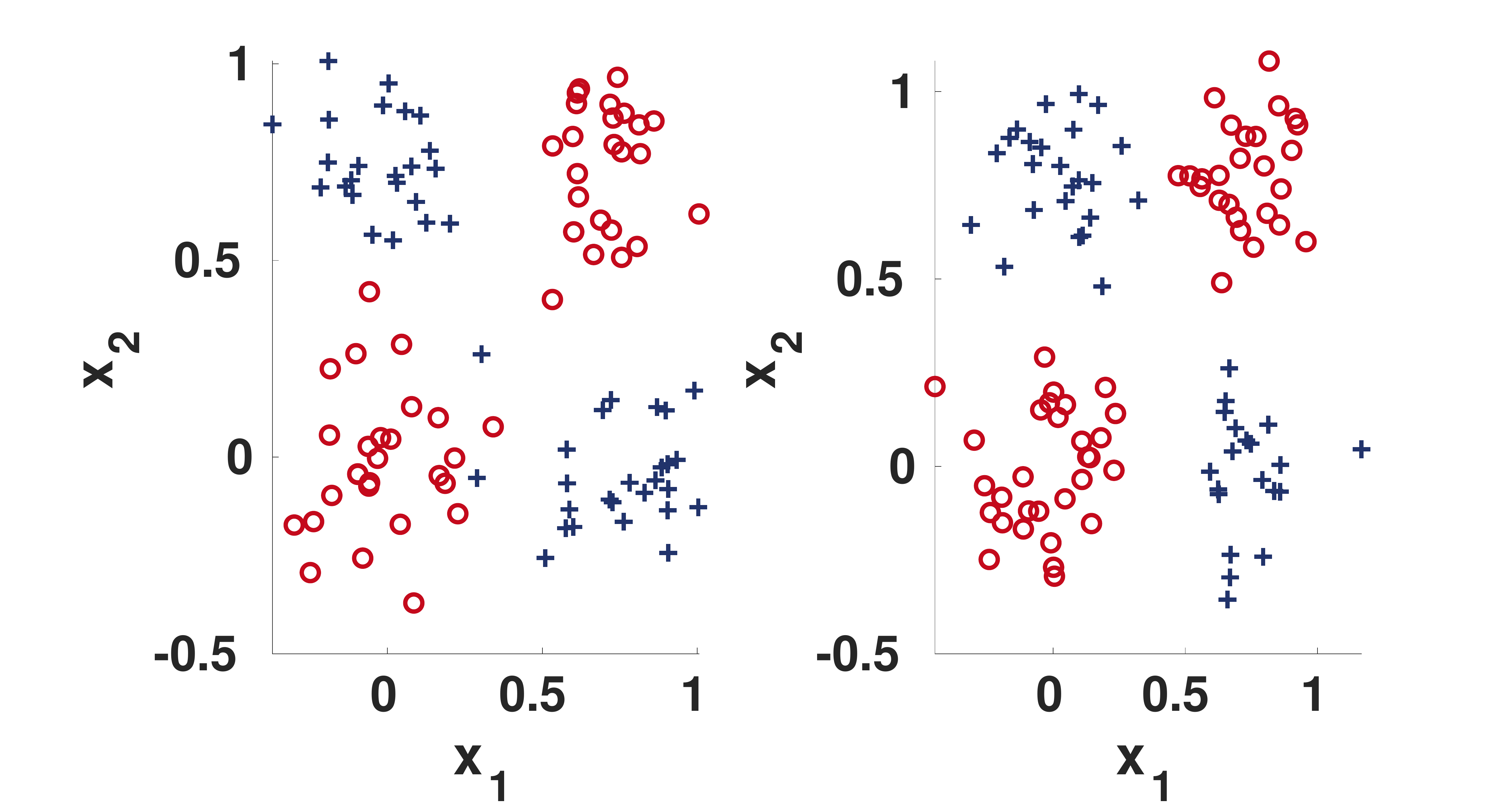}
		\caption{synthetic two class xor train and test data}
		\label{svm_data_fig}	
	\end{subfigure}\qquad
	\begin{subfigure}[t]{0.3\textwidth}
		
		\includegraphics[trim=20 0 0 0,clip,width=1\linewidth]{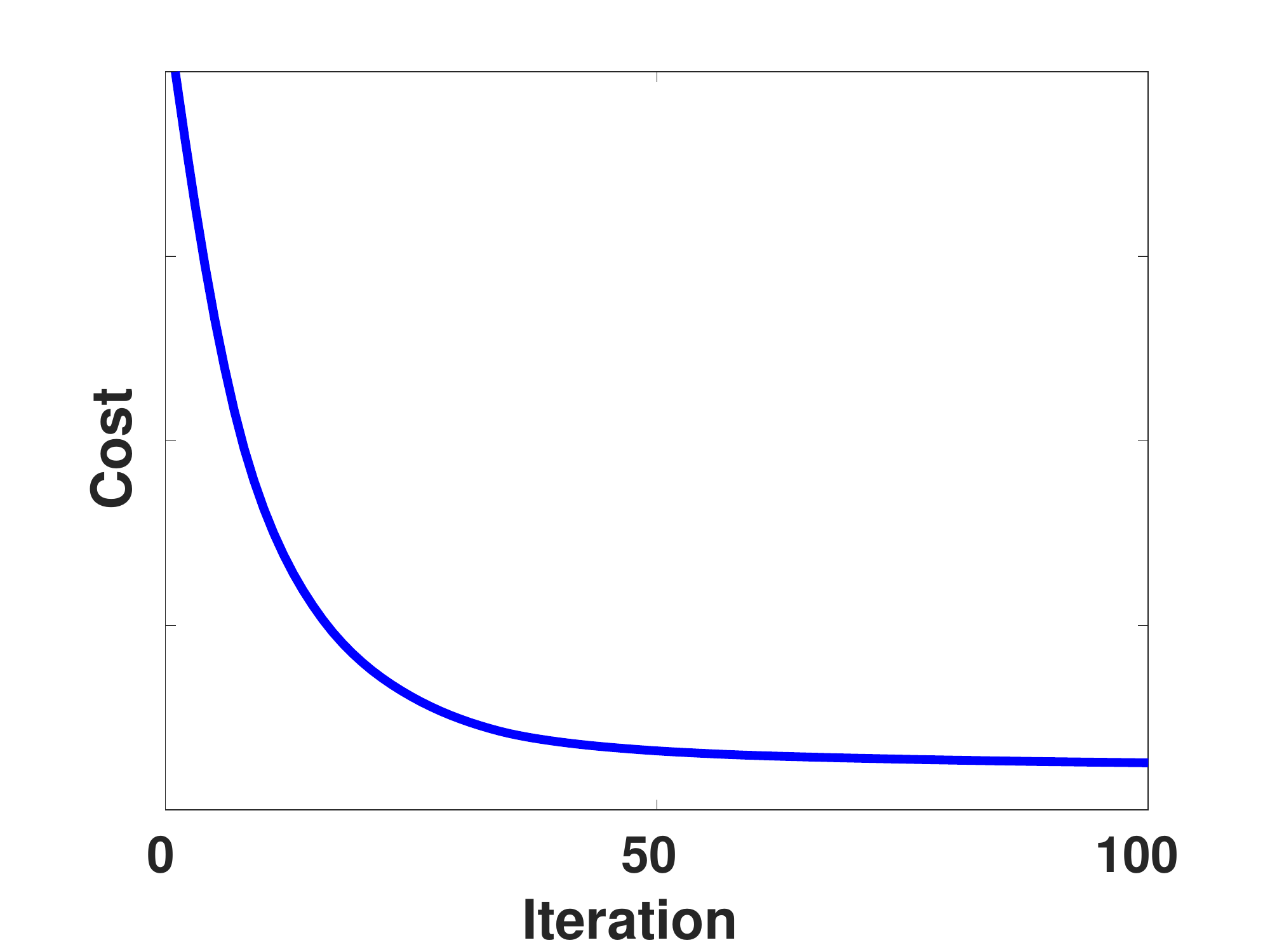}
		\caption{SVM training curve}
		\label{svm_train_fig}	
	\end{subfigure}%
	\begin{subfigure}[t]{0.3\textwidth}
		\centering
		\includegraphics[trim=20 0 0 0,clip,width=1.1\linewidth,height=4.1cm]{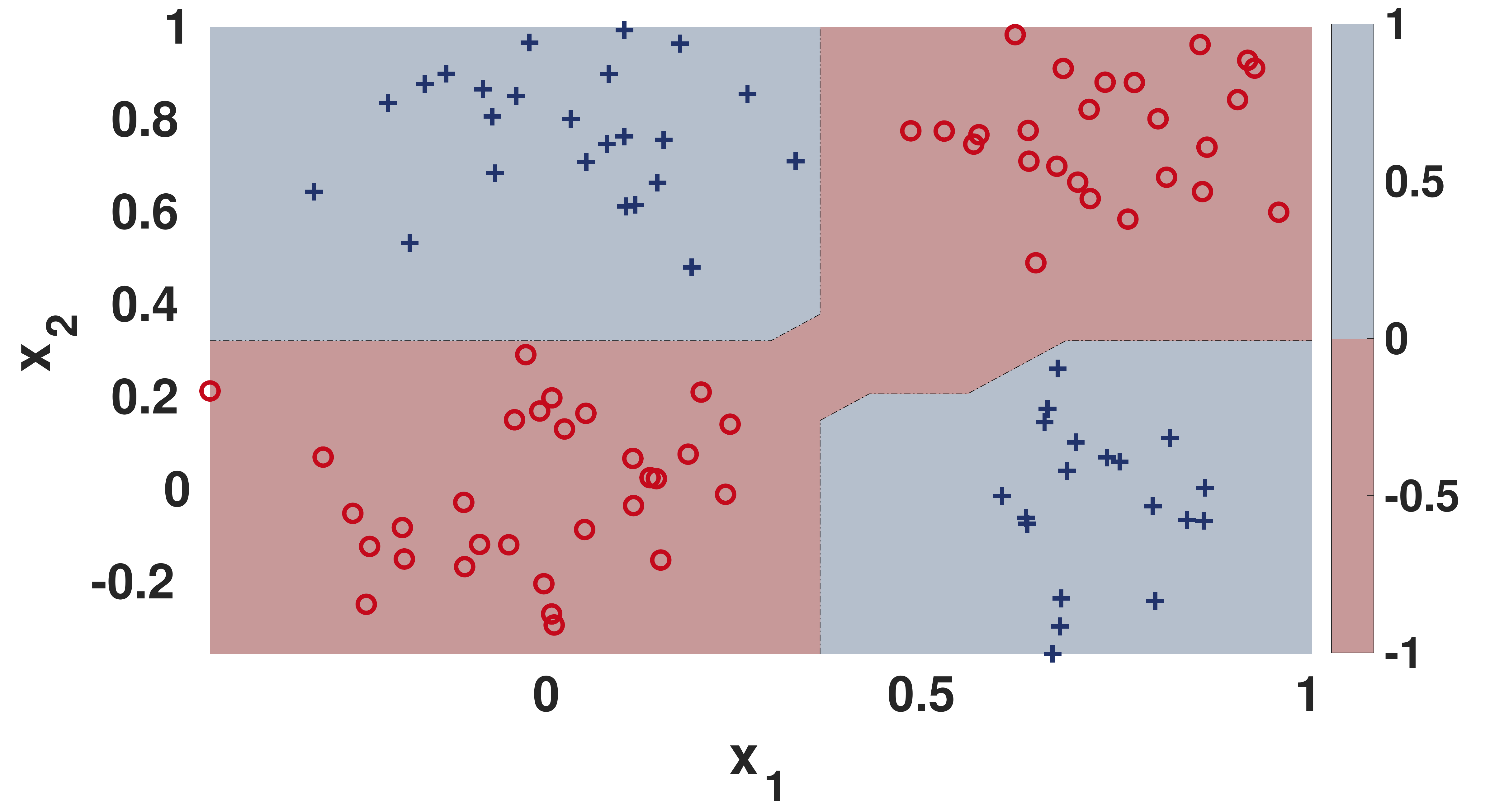}
		\caption{Decision boundary plot of MP based SVM classification result}
		\label{svm_cont_fig}
	\end{subfigure}%
	%

	\caption{}
	
\end{figure*}

We also implement a support vector machine (SVM) using the MP algorithm for a two class non linearly separable problem. We use a Cauchy kernel and by choosing the appropriate normalization, parameters are converted into positive to apply MP approximation.

The formulation is as follows;

For a given input $\tilde{x}$,
\begin{equation}
f(\tilde{x})= \sum_{s}^{}\tilde{w_s}\tilde{K}(\tilde{x_s},\tilde{x})= \sum_{s}^{}(\tilde{w_s}^+-\tilde{w_s}^-)(\tilde{K_s}^+-\tilde{K_s}^-)
\label{smv_eq1}
\end{equation}
where $\tilde{K}$ is the kernel function, $\tilde{x_s}$ is the $s^{th} $ support vector and $\tilde{x}$ is the input sample (Here $\tilde{.}$ indicates that the parameters are not in the log likelihood domain).
\begin{align}
f^+-f^-= \left(\sum_{s}^{}\tilde{w_s}^+\tilde{K_s}^+ + \tilde{w_s}^-\tilde{K_s}^-\right)  \nonumber \\- \left(\sum_{s}^{}\tilde{w_s}^+\tilde{K_s}^- +\tilde{w_s}^-\tilde{K_s}^+\right)
\label{smv_eq2}
\end{align}
Converting into log likelihood domain
\begin{align}
L_f^+ - L_f^- = log\left( \sum_{s}^{}e^{w_s^+ +K_s^+} + e^{w_s^- + K_s^-}\right) \nonumber \\  -  log\left( \sum_{s}^{}e^{w_s^+ +K_s^-} + e^{w_s^- + K_s^+}\right)
\label{svm_eq3}
\end{align}
The above can be approximated using MP algorithm as;
\begin{align}
L_f^+ - L_f^- = MP\left( \{w_s^+ +K_s^+, w_s^- + K_s^-\}, \gamma \right) \nonumber \\  -  MP\left( \{w_s^+ +K_s^-, w_s^- + K_s^+ \}, \gamma\right)
\label{svm_eq4}
\end{align}
The formulation in eq. \eqref{svm_eq4} is similar to that in eq.s \eqref{eq3_per} and \eqref{eq4_per}. Hence the parameter update rules are similar to that of perceptron defined in sec. \ref{perc_paraupdate} using $l_1$ norm \eqref{eq1_per1} as the cost function.

\subsection{Kernel function}
\begin{equation}
K(\tilde{x}_s,\tilde{x}) = K_s = K_s^+-K_s^-
\end{equation}
We use a Cauchy kernel function given as,
\begin{equation}
\tilde{K}(\tilde{x_s},\tilde{x})= \frac{1}{c+||\tilde{x_s}-\tilde{x}||_2^2}
\label{svm_eq5}
\end{equation}
Ensuring $|\tilde{x}| <1 $ or $\tilde{x}= \tilde{x}^+-\tilde{x}^-$ such that $ \tilde{x}^++\tilde{x}^- =1$
we get,
\begin{align}
\sum_{i}^{}(\tilde{x}_{si}-\tilde{x}_i)^2= \sum_{i}^{}\left(\tilde{x}_{si}^+ + \tilde{x}_i^- - \tilde{x}_{si}^- -\tilde{x}_i^+\right)^2 \nonumber \\ 
= \sum_{i}^{}(\tilde{x}_{si}^{+^2} +  \tilde{x}_{si}^{-^2} + \tilde{x}_i^{+^2} +  \tilde{x}_i^{-^2} - 2\tilde{x}_{si}^{+}\tilde{x}_{si}^{-} + 2\tilde{x}_{si}^{+}\tilde{x}_i^{-} \nonumber \\ 
- 2\tilde{x}_{si}^{+}\tilde{x}_i^{+} + 2\tilde{x}_{si}^{-}\tilde{x}_i^{+} -2\tilde{x}_{si}^{-}\tilde{x}_i^{-}  -2\tilde{x}_i^{+}\tilde{x}_i^{-}) \label{svm_eq6}
\end{align}
Here $\tilde{x_{si}}$ indicates the $i^{th}$ sample of the support vector $\tilde{x_{s}}$

Each of the terms in eq. \ref{svm_eq6} is added with a constant $c$ as per eq. \eqref{svm_eq5}. By choosing appropriate value for $c$, parameters can be converted to positive values to apply MP approximation as given below;

Consider the negative term

 \begin{align}
-2\tilde{x}_{si}^{+}\tilde{x}_{si}^{-} + 2 \implies -2\tilde{x}_{si}^{+}\tilde{x}_{si}^{-} + 2(\tilde{x}_{si}^{+} + \tilde{x}_{si}^-) \nonumber \\
\implies 2\tilde{x}_{si}^{-} + 2\tilde{x}_{si}^{+}(1-\tilde{x}_{si}^{-}) \nonumber \\
\implies 2\tilde{x}_{si}^{-} + 2\tilde{x}_{si}^{+^2}
\label{svm_eq7}
 \end{align}
 which ensures all such terms to be positive and hence MP approximation can be applied.
 
 Converting into the log-likelihood domain
 \begin{align}
 log\left[\frac{1}{C+\sum_{i}^{}(\tilde{x}_{si}-\tilde{x}_i)^2} \right]=  -log \left[\sum_{i}^{}\tilde{x}_{si}^+\tilde{x}_{si}^+ + \tilde{x}_{si}^+\tilde{x}_i^- + \dots \right] \nonumber \\
  = -log \left[\sum_{i}^{}e^{x_{si}^+ +x_{si}^+} + e^{x_{si}^+ + x_i^- +} \dots \right] \nonumber \\ 
 \label{svm_eq8}
 \end{align}
 Hence
 
 \begin{equation}
 K_s = MP \left( \{ x_{si}^+ + x_{si}^+, x_{si}^+ + x_i^-, \dots\}, \gamma_2 \right)
 \label{svm_eq9}
 \end{equation}
 \begin{table}[htbp]
 	\resizebox{0.5\textwidth}{!}{
 		\begin{tabular}{|c|c|c|c|c|c|c|}
 			\hline
 			& \multicolumn{3}{c|}{\textbf{Train}}  & \multicolumn{3}{c|}{\textbf{Test}}   \\ \hline
 			& Class 1 & Class 2 & \textbf{Overall}  & Class 1 & Class 2 & \textbf{Overall} \\ \hline
 			Accuracy (\%)       & $99$       & $96$    &$97$          & $100$       & $100$    & $100$    \\ \hline
 		\end{tabular}
 	}
 	\caption{Classification accuracies on synthetic Xor data for MP SVM }
 	\label{tab_svm}
 \end{table}
 \begin{table*}[htbp]
 	\centering
 	\resizebox{0.9\textwidth}{!}{  
 		\begin{tabular}{|c|c|c|c|c|c|}
 			\hline
 			& \multicolumn{2}{c|}{\textbf{Conventional SVM}}  & \multicolumn{2}{c|}{\textbf{MP SVM}}   \\ \hline
 			& Train Acc & Test Acc & Train Acc & Test Acc \\ \hline
 			AReM Bending : dim 7               & $96\%$   & $95\%$   & $95 \%$       & $94\%$  \\ \hline
 			AReM Lying : dim 7             & $96 \%$   & $90\% $    & $95 \%$       & $94\%$  \\ \hline
 			
 			Wisconsin-breast-cancer : dim 10             & $98\%$   & $98\%$     & $97\%$       & $98\%$  \\ \hline
 			
 			Statlog Heart : dim 13              & $87 \%$   & $81\% $     & $83 \%$       & $81\%$ \\ \hline
 			Pima Indians diabetes : dim 8               & $81 \%$   & $73\% $  & $74 \%$       & $74\%$   \\ \hline
 		\end{tabular}
 	}
 	\caption{Classification accuracies on UCI dataset using MP based SVM}
 	\label{tab_uci_svm}
 \end{table*}
 
\subsection{Implementation and results}
\subsubsection{Synthetic Xor data}
For evaluating our SVM formulation explained in sec.\ref{svm}, we use  a synthetic non-linearly separable xor data for training and inference using MATLAB. In this case also we use 100 train and test samples each. 
The scatter plot of the dataset is shown in Fig. \ref{svm_data_fig}.  Figure \ref{svm_train_fig} shows the training curve showing a decreasing cost function during each iteration. The classification accuracies for the train and test data shown in Table \ref{tab_svm} and the decision boundary plot  of the inference results  in Fig. \ref{svm_cont_fig} show the effectiveness of the algorithm as a non-linear classifier.

 \subsubsection{UCI  datasets}
  
 The performances of MP based SVM on different benchmark UCI  datasets  such as Stalog  Heart,  Diabetes, Wisconsin-breast-cancer and Activity recognition (AReM) datasets \cite{blake1998uci} is also evaluated and the results are compared with  a conventional SVM using Cauchy's kernel. Table \ref{tab_uci_svm} shows the performances in terms of classification accuracy (acc). The dimension (dim) of each dataset is also shown in the table.  It can be seen that MP based SVM also gives  performances comparable to that of conventional SVM with the added benefit of significant reduction in computational complexity and improvement in energy cost as discussed in sec. \ref{complexity1}.

\subsection{Complexity}
During the inference stage, for a given sample $x$ of dimension $N$, the output for conventional SVM is given as,

\begin{equation}
f(x)= \sum_{s}^{S}w_s K(x_s,x)
\end{equation}
where, $S>N$ is the total number of support vectors, $K$ is the  Kernel function and $x_s$  is the $s^{th}$ support vector.  For this MVM operation, the overall complexity is given as,
\begin{align}
C_{SVM-C}= S \times C_M+ S \times C_A\\
\end{align}
where $C_{SVM-C}$ is the complexity of inference in conventional SVM. $C_M$ and $C_A$ are the complexity of multiplication and addition operations respectively.

For MP based SVM, from sec.\ref{svm} equation \eqref{svm_eq4} we get,

\begin{align}
L_f^+ - L_f^- = MP\left( \{w_s^+ +K_s^+, w_s^- + K_s^-\}, \gamma \right) \nonumber \\  -  MP\left( \{w_s^+ +K_s^-, w_s^- + K_s^+ \}, \gamma\right)
\label{svm_eq4dupe}
\end{align}

Considering the term $L_f^+$,the complexity is given as,
\begin{align}
C_{SVM-MP}=  2\times S \times C_A + F \times log(2\times S) \times C_c\\
\end{align}
where $F$ is the sparsity factor of the thresholding operation determined by $\gamma$, $C_A$ is the complexity of addition and $C_c$ is the complexity of comparison operation. As explained in \secref{complexity1}, $C_A = 3 \times C_S$ where shift is considered as an elementary operation and 2 complete multiplications require $d^2$ full adders, where $d$ is the number of bits as given in \cite{energypaper}.

\begin{figure}[h!]%
	
	
		\includegraphics[width=1\linewidth]{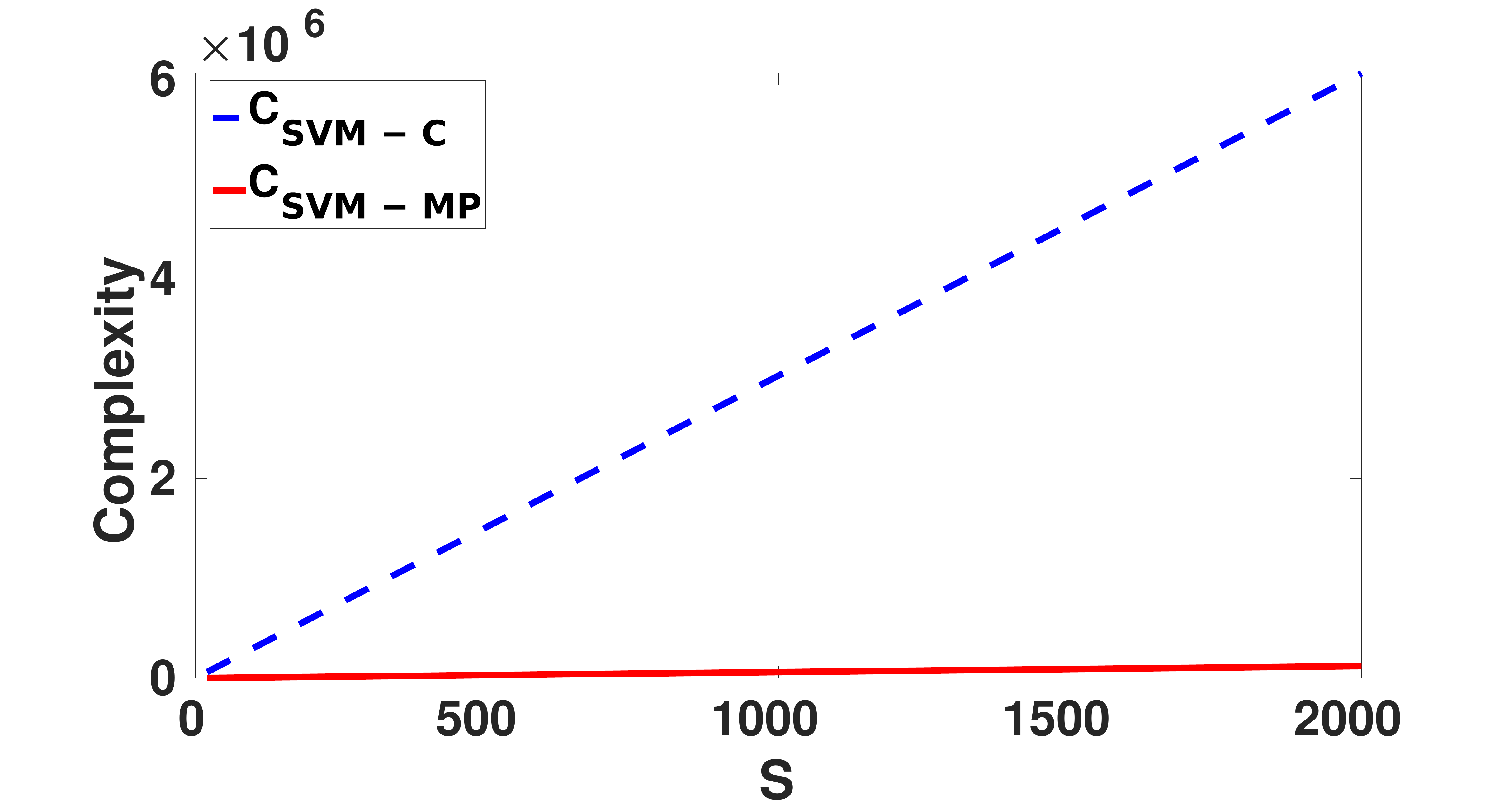}
		\caption{$C_{SVM-C}$ and $C_{SVM-MP}$ plots w.r.t S for various datasets}
		\label{svm_comp_plot}	

	
\end{figure}

Figure \ref{svm_comp_plot} shows how the complexities $C_{SVM-C}$ and $C_{SVM-MP}$ varies with $S$. We assume a 10 bit multiplication, addition and comparison for computing the complexities with varying values of $S$ depending on various datasets. The sparsity factor $K$ is assumed to be 1, which can be further reduced by adjusting the parameter $\gamma$.
It can be inferred from the plot that the inference complexity of MP-SVM is significantly lower than that of conventional SVM.

As discussed in sec.\ref{complexity1}, this will result in  significant improvement of energy cost. Similar to MP perceptron and MP MLP,  the cost function  used for MP SVM is $L_1$ which in conjunction with ReLU operation ensures network sparsity.

\section{Conclusion}\label{conc}
 In this paper we proposed an alternate hardware-software codesign of ML and neural network architectures. The architecture only uses simple addition and thresholding operations to implement inference and learning instead of using MVM operations and non-linear activation functions. The margin-propagation based computation maps multiplications into additions and additions into a dynamic rectifying-linear-unit (ReLU) operation which results in significant improvement in computational and hence energy cost. The formulation also enables network sparsity. We showed the application of MP formulation for the design of linear classifiers, shallow multi-layer perceptrons as well as  support vector machines and evaluated the performance of the same on synthetic data and UCI benchmark database. The algorithms gave comparable performances to that of their conventional counterparts showing their usefulness for IoT and edge ML platforms.  

\appendices

\section{MP-Perceptron training:  Proof for error-function derivatives}\label{appendixa}
Considering a two class problem class$+$ and class$-$, the error function can be written as;
\begin{equation}
E= \sum_{n}^{}|y_{n}^{+}-p^+| + |y_{n}^{-}-p^-|
\label{eq1_per}
\end{equation}
where 

$y_{n}^{+}$: label for class$+$ for $n^{th}$ sample \\

$y_{n}^{-}$: label for class$-$ for $n^{th}$ sample\\

$y_{n}^{+} + y_{n}^{-} =1$


From eq. \eqref{eq1_per}
\begin{equation}
\frac{\partial E}{\partial w_{i}^+} = \sum_{n}^{}sign(p^+ - y_{n}^+)\frac{\partial p^+}{\partial w_{i}^+} + sign(p^- - y_{n}^-)\frac{\partial p^-}{\partial w_{i}^+}
\label{eq5_per2}
\end{equation}
If $\bar{x}$ is the input to the MP algorithm such that,
$z=MP(\bar{x},\gamma) \implies \sum_{d}^{}[x_{d}-z]_+$ where $d$ indicates each element of $\bar{x}$ then,

\begin{equation}
\frac{\partial z}{\partial x_{d}} = \frac{1}{\textit{A} }\mathbbm{1}(x_{d}>z)
\label{eq6_per}
\end{equation}
where $\textit{A} $ indicates the number of $x_d$ such that $x_{d} > z$ and $\mathbbm{1}$ is the indicator function.
Also

\begin{equation}
\frac{\partial [x_{d}-z]_+}{\partial x_{d}} = \left(1-\frac{1}{\textit{A} }\right)\mathbbm{1}(x_{d}>z)
\label{eq7_per}
\end{equation}

Using equations \eqref{eq2_per}, \eqref{eq3_per}, \eqref{eq6_per} and \eqref{eq7_per}
\begin{equation}
\frac{\partial p^+}{\partial w_{i}^+}= \left(1-\frac{1}{\textit{A} }\right)\mathbbm{1}(z^+>z) \frac{1}{\textit{A} p}\mathbbm{1}(x_{i}^+ + w_{i}^+ >z^+)
\label{eq8_per2}
\end{equation}
Similarly using \eqref{eq2_per}, \eqref{eq4_per}, \eqref{eq6_per} and \eqref{eq7_per}
\begin{equation}
\frac{\partial p^-}{\partial w_{i}^+}= \left(1-\frac{1}{\textit{A} }\right)\mathbbm{1}(z^->z) \frac{1}{\textit{A} n}\mathbbm{1}(x_{i}^- + w_{i}^+ >z^-)
\label{eq9_per}
\end{equation}
Substituting \eqref{eq8_per2} and \eqref{eq9_per} in \eqref{eq5_per2} we get, $\frac{\partial E}{\partial w_{i}^+}$

Similarly,
\begin{equation}
\frac{\partial E}{\partial w_{i}^-} = \sum_{n}^{}sign(p^+ - y_{n}^+)\frac{\partial p^+}{\partial w_{i}^-} + sign(p^- - y_{n}^-)\frac{\partial p^-}{\partial w_{i}^-}
\label{eq10_per}
\end{equation}
where,

\begin{equation}
\frac{\partial p^+}{\partial w_{i}^-}= \left(1-\frac{1}{\textit{A} }\right)\mathbbm{1}(z^+>z) \frac{1}{\textit{A} p}\mathbbm{1}(x_{i}^- + w_{i}^- >z^+)
\label{eq11_per}
\end{equation}

\begin{equation}
\frac{\partial p^-}{\partial w_{i}^-}= \left(1-\frac{1}{\textit{A} }\right)\mathbbm{1}(z^->z) \frac{1}{\textit{A} n}\mathbbm{1}(x_{i}^+ + w_{i}^- >z^-)
\label{eq12_per}
\end{equation}

\subsection{Derivatives with respect to bias}
From eq. \eqref{eq1_per}
\begin{equation}
\frac{\partial E}{\partial b^+} = \sum_{n}^{}sign(p^+ - y_{n}^+)\frac{\partial p^+}{\partial b^+} 
\label{eq5_bias_per}
\end{equation}
As $\frac{\partial p^-}{\partial b^+}=0$

Using equations \eqref{eq2_per}, \eqref{eq3_per}, \eqref{eq6_per} and \eqref{eq7_per}
\begin{equation}
\frac{\partial p^+}{\partial b^+}= \left(1-\frac{1}{\textit{A} }\right)\mathbbm{1}(z^+>z) \frac{1}{\textit{A} p}\mathbbm{1}(b^+ >z^+)
\label{eq8_bias_per}
\end{equation}

Similarly,
\begin{equation}
\frac{\partial E}{\partial b^-} = \sum_{n}^{}sign(p^- - y_{n}^-)\frac{\partial p^-}{\partial b^-}
\label{eq10_bias_per}
\end{equation}
Using \eqref{eq2_per}, \eqref{eq4_per}, \eqref{eq6_per} and \eqref{eq7_per}

\begin{equation}
\frac{\partial p^-}{\partial b^-}= \left(1-\frac{1}{\textit{A} }\right)\mathbbm{1}(z^->z) \frac{1}{\textit{A} n}\mathbbm{1}(b^- >z^-)
\label{eq12_bias_per}
\end{equation}

Using the error gradients obtained from above, the weights and bias are updated during each iteration as follows;
\begin{equation}
{}_{}w_{i,(\tau)}^{+}= {}_{}w_{i,(\tau-1)}^{+} - \epsilon \frac{\partial E}{\partial w_{i,(\tau-1)}^+} \\
\end{equation}
\begin{equation}
{}_{}w_{i,(\tau)}^{-}= {}_{}w_{i,(\tau-1)}^{-} - \epsilon \frac{\partial E}{\partial w_{i,(\tau-1)}^-} \\
\end{equation}
\begin{equation}
{}_{}b_{(\tau)}^{+}= {}_{}b_{(\tau-1)}^{+} - \epsilon \frac{\partial E}{\partial b_{(\tau-1)}^+} \\
\end{equation}
\begin{equation}
{}_{}b_{(\tau)}^{-}= {}_{}b_{(\tau-1)}^{-} - \epsilon \frac{\partial E}{\partial b_{(\tau-1)}^-} \\
\end{equation}
where $\epsilon$ is the learning rate and $\tau$ indicates the iteration step.

\section{MP-MLP training: Proof for error-function derivatives}\label{appendixb}
Considering a two class problem class$+$ and class$-$, the error function can be written as;
\begin{equation}
E= \sum_{n}^{}|y_{nk}^{+}-p_{k}^+| + |y_{nk}^{-}-p_{k}^-|
\label{eq1}
\end{equation}
where 

$y_{nk}^{+}$: label for class$+$ for $n^{th}$ sample \\

$y_{nk}^{-}$: label for class$-$ for $n^{th}$ sample\\

$y_{nk}^{+} + y_{nk}^{-} =1$

\subsection*{Output layer $K$}

From eq. \eqref{eq1}
\begin{equation}
\frac{\partial E}{\partial w_{jk}^+} = \sum_{n}^{}sign(p_{k}^+ - y_{nk}^+)\frac{\partial p_{k}^+}{\partial w_{jk}^+} + sign(p_{k}^- - y_{nk}^-)\frac{\partial p_{k}^-}{\partial w_{jk}^+}
\label{eq5}
\end{equation}
%
%
%

Using equations \eqref{eq2}, \eqref{eq3}, \eqref{eq6_per} and \eqref{eq7_per}
\begin{equation}
\frac{\partial p_{k}^+}{\partial w_{jk}^+}= \left(1-\frac{1}{\textit{A} _k}\right)\mathbbm{1}(z_{k}^+>z_{k}) \frac{1}{\textit{A} p_k}\mathbbm{1}(p_{j}^+ + w_{jk}^+ >z_{k}^+)
\label{eq8}
\end{equation}
Similarly using \eqref{eq2}, \eqref{eq4}, \eqref{eq6_per} and \eqref{eq7_per}
\begin{equation}
\frac{\partial p_{k}^-}{\partial w_{jk}^+}= \left(1-\frac{1}{\textit{A} _k}\right)\mathbbm{1}(z_{k}^->z_{k}) \frac{1}{\textit{A} n_k}\mathbbm{1}(p_{j}^- + w_{jk}^+ >z_{k}^-)
\label{eq9}
\end{equation}
Substituting \eqref{eq8} and \eqref{eq9} in \eqref{eq5} we get, $\frac{\partial E}{\partial w_{jk}^+}$

Similarly,
\begin{equation}
\frac{\partial E}{\partial w_{jk}^-} = \sum_{n}^{}sign(p_{k}^+ - y_{nk}^+)\frac{\partial p_{k}^+}{\partial w_{jk}^-} + sign(p_{k}^- - y_{nk}^-)\frac{\partial p_{k}^-}{\partial w_{jk}^-}
\label{eq10}
\end{equation}
where,

\begin{equation}
\frac{\partial p_{k}^+}{\partial w_{jk}^-}= \left(1-\frac{1}{\textit{A} _k}\right)\mathbbm{1}(z_{k}^+>z_{k}) \frac{1}{\textit{A} p_k}\mathbbm{1}(p_{j}^- + w_{jk}^- >z_{k}^+)
\label{eq11}
\end{equation}

\begin{equation}
\frac{\partial p_{k}^-}{\partial w_{jk}^-}= \left(1-\frac{1}{\textit{A} _k}\right)\mathbbm{1}(z_{k}^->z_{k}) \frac{1}{\textit{A} n_k}\mathbbm{1}(p_{j}^+ + w_{jk}^- >z_{k}^-)
\label{eq12}
\end{equation}

\subsection{Derivatives with respect to bias}
From eq. \eqref{eq1}
\begin{equation}
\frac{\partial E}{\partial b_{k}^+} = \sum_{n}^{}sign(p_{k}^+ - y_{nk}^+)\frac{\partial p_{k}^+}{\partial b_{k}^+} 
\label{eq5_bias}
\end{equation}
As $\frac{\partial p_{k}^-}{\partial b_{k}^+}=0$

Using equations \eqref{eq2}, \eqref{eq3}, \eqref{eq6_per} and \eqref{eq7_per}
\begin{equation}
\frac{\partial p_{k}^+}{\partial b_{k}^+}= \left(1-\frac{1}{\textit{A} _k}\right)\mathbbm{1}(z_{k}^+>z_{k}) \frac{1}{\textit{A} p_k}\mathbbm{1}(b_{k}^+ >z_{k}^+)
\label{eq8_bias}
\end{equation}

Similarly,
\begin{equation}
\frac{\partial E}{\partial b_{k}^-} = \sum_{n}^{}sign(p_{k}^- - y_{nk}^-)\frac{\partial p_{k}^-}{\partial b_{k}^-}
\label{eq10_bias}
\end{equation}
Using \eqref{eq2}, \eqref{eq4}, \eqref{eq6_per} and \eqref{eq7_per}

\begin{equation}
\frac{\partial p_{k}^-}{\partial b_{k}^-}= \left(1-\frac{1}{\textit{A} _k}\right)\mathbbm{1}(z_{k}^->z_{k}) \frac{1}{\textit{A} n_k}\mathbbm{1}(b_{k}^- >z_{k}^-)
\label{eq12_bias}
\end{equation}

\subsection*{Hidden layer $J$}

From \eqref{eq1}
\begin{equation}
\frac{\partial E}{\partial w_{ij}^+} = \sum_{n}^{}sign(p_{k}^+ - y_{nk}^+)\frac{\partial p_{k}^+}{\partial w_{ij}^+} + sign(p_{k}^- - y_{nk}^-)\frac{\partial p_{k}^-}{\partial w_{ij}^+}
\label{eq16}
\end{equation}
Using equations \eqref{eq2},\eqref{eq3}, \eqref{eq13}, \eqref{eq14} and \eqref{eq15} we get,
\begin{equation}
\frac{\partial p_{k}^+}{\partial w_{ij}^+} = \frac{\partial p_{k}^+}{\partial z_{k}^+}\frac{\partial z_{k}^+}{\partial p_{j}^+}\frac{\partial p_{j}^+}{\partial w_{ij}^+} +  \frac{\partial p_{k}^+}{\partial z_{k}^+}\frac{\partial z_{k}^+}{\partial p_{j}^-}\frac{\partial p_{j}^-}{\partial w_{ij}^+} \nonumber
\end{equation}
\begin{align}
= \left(1-\frac{1}{\textit{A} _k}\right)\mathbbm{1}(z_{k}^+>z_{k})
\frac{1}{\textit{A} p_k}\mathbbm{1}(p_{j}^+ + w_{jk}^+ >z_{k}^+) \nonumber\\
\left(1-\frac{1}{\textit{A} _j}\right)\mathbbm{1}(z_{j}^+>z_{j})
\frac{1}{\textit{A} p_j}\mathbbm{1}(x_{i}^+ + w_{ij}^+ >z_{j}^+) \nonumber\\
+  \left(1-\frac{1}{\textit{A} _k}\right)\mathbbm{1}(z_{k}^+>z_{k})
\frac{1}{\textit{A} p_k}\mathbbm{1}(p_{j}^- + w_{jk}^- >z_{k}^+) \nonumber\\
\left(1-\frac{1}{\textit{A} _j}\right)\mathbbm{1}(z_{j}^->z_{j})
\frac{1}{\textit{A} n_j}\mathbbm{1}(x_{i}^- + w_{ij}^+ >z_{j}^-)\nonumber\\
\label{eq17}
\end{align}

Using equations \eqref{eq2},\eqref{eq4}, \eqref{eq13}, \eqref{eq14} and \eqref{eq15} we get,

\begin{flalign*}
\frac{\partial p_{k}^-}{\partial w_{ij}^+} &= & \\
\end{flalign*}
\begin{align}
\left(1-\frac{1}{\textit{A} _k}\right)\mathbbm{1}(z_{k}^->z_{k})
\frac{1}{\textit{A} n_k}\mathbbm{1}(p_{j}^- + w_{jk}^+ >z_{k}^-) \nonumber\\
\left(1-\frac{1}{\textit{A} _j}\right)\mathbbm{1}(z_{j}^->z_{j})
\frac{1}{\textit{A} n_j}\mathbbm{1}(x_{i}^- + w_{ij}^+ >z_{j}^-)\nonumber\\
+  \left(1-\frac{1}{\textit{A} _k}\right)\mathbbm{1}(z_{k}^->z_{k})
\frac{1}{\textit{A} n_k}\mathbbm{1}(p_{j}^+ + w_{jk}^- >z_{k}^-) \nonumber\\
\left(1-\frac{1}{\textit{A} _j}\right)\mathbbm{1}(z_{j}^+>z_{j})
\frac{1}{\textit{A} p_j}\mathbbm{1}(x_{i}^+ + w_{ij}^+ >z_{j}^+) \nonumber\\
\label{eq19}
\end{align}
Similarly,
\begin{equation}
\frac{\partial E}{\partial w_{ij}^-} = \sum_{n}^{}sign(p_{k}^+ - y_{nk}^+)\frac{\partial p_{k}^+}{\partial w_{ij}^-} + sign(p_{k}^- - y_{nk}^-)\frac{\partial p_{k}^-}{\partial w_{ij}^-}
\label{eq20}
\end{equation}
where,

\begin{flalign*}
\frac{\partial p_{k}^+}{\partial w_{ij}^-} &= & \\
\end{flalign*}
\begin{align}
\left(1-\frac{1}{\textit{A} _k}\right)\mathbbm{1}(z_{k}^+>z_{k})
\frac{1}{\textit{A} p_k}\mathbbm{1}(p_{j}^+ + w_{jk}^+ >z_{k}^+) \nonumber\\
\left(1-\frac{1}{\textit{A} _j}\right)\mathbbm{1}(z_{j}^+>z_{j})
\frac{1}{\textit{A} p_j}\mathbbm{1}(x_{i}^- + w_{ij}^- >z_{j}^+) \nonumber\\
+  \left(1-\frac{1}{\textit{A} _k}\right)\mathbbm{1}(z_{k}^+>z_{k})
\frac{1}{\textit{A} p_k}\mathbbm{1}(p_{j}^- + w_{jk}^- >z_{k}^+) \nonumber\\
\left(1-\frac{1}{\textit{A} _j}\right)\mathbbm{1}(z_{j}^->z_{j})
\frac{1}{\textit{A} n_j}\mathbbm{1}(x_{i}^+ + w_{ij}^- >z_{j}^-)\nonumber\\
\label{eq21}
\end{align}
and
\begin{flalign*}
\frac{\partial p_{k}^-}{\partial w_{ij}^-} &= & \\
\end{flalign*}
\begin{align}
\left(1-\frac{1}{\textit{A} _k}\right)\mathbbm{1}(z_{k}^->z_{k})
\frac{1}{\textit{A} n_k}\mathbbm{1}(p_{j}^- + w_{jk}^+ >z_{k}^-) \nonumber\\
\left(1-\frac{1}{\textit{A} _j}\right)\mathbbm{1}(z_{j}^->z_{j})
\frac{1}{\textit{A} n_j}\mathbbm{1}(x_{i}^+ + w_{ij}^- >z_{j}^-)\nonumber\\
+  \left(1-\frac{1}{\textit{A} _k}\right)\mathbbm{1}(z_{k}^->z_{k})
\frac{1}{\textit{A} n_k}\mathbbm{1}(p_{j}^+ + w_{jk}^- >z_{k}^-) \nonumber\\
\left(1-\frac{1}{\textit{A} _j}\right)\mathbbm{1}(z_{j}^+>z_{j})
\frac{1}{\textit{A} p_j}\mathbbm{1}(x_{i}^- + w_{ij}^- >z_{j}^+) \nonumber\\
\label{eq22}
\end{align}

\subsubsection{Derivatives with respect to bias}
From \eqref{eq1}
\begin{equation}
\frac{\partial E}{\partial b_{j}^+} = \sum_{n}^{}sign(p_{k}^+ - y_{nk}^+)\frac{\partial p_{k}^+}{\partial b_{j}^+} + sign(p_{k}^- - y_{nk}^-)\frac{\partial p_{k}^-}{\partial b_{j}^+}
\label{eq16_bias}
\end{equation}
Using equations \eqref{eq2},\eqref{eq3}, \eqref{eq13}, \eqref{eq14} and \eqref{eq15} we get,
\begin{flalign*}
\frac{\partial p_{k}^+}{\partial b_{j}^+} = \frac{\partial p_{k}^+}{\partial z_{k}^+}\frac{\partial z_{k}^+}{\partial p_{j}^+}\frac{\partial p_{j}^+}{\partial b_{j}^+} +  0 \nonumber
\end{flalign*}
\begin{align}
= \left(1-\frac{1}{\textit{A} _k}\right)\mathbbm{1}(z_{k}^+>z_{k})
\frac{1}{\textit{A} p_k}\mathbbm{1}(p_{j}^+ + w_{jk}^+ >z_{k}^+) \nonumber\\
\left(1-\frac{1}{\textit{A} _j}\right)\mathbbm{1}(z_{j}^+>z_{j})
\frac{1}{\textit{A} p_j}\mathbbm{1}(b_{j}^+ >z_{j}^+) \nonumber\\
\label{eq17_bias}
\end{align}

Using equations \eqref{eq2},\eqref{eq4}, \eqref{eq13}, \eqref{eq14} and \eqref{eq15} we get,

\begin{flalign*}
\frac{\partial p_{k}^-}{\partial b_{j}^+} &= & \\
\end{flalign*}
\begin{align}
\left(1-\frac{1}{\textit{A} _k}\right)\mathbbm{1}(z_{k}^->z_{k})
\frac{1}{\textit{A} n_k}\mathbbm{1}(p_{j}^+ + w_{jk}^- >z_{k}^-) \nonumber\\
\left(1-\frac{1}{\textit{A} _j}\right)\mathbbm{1}(z_{j}^+>z_{j})
\frac{1}{\textit{A} p_j}\mathbbm{1}(b_{j}^+ >z_{j}^+) \nonumber\\
\label{eq19_bias}
\end{align}

Similarly,
\begin{equation}
\frac{\partial E}{\partial b_{j}^-} = \sum_{n}^{}sign(p_{k}^+ - y_{nk}^+)\frac{\partial p_{k}^+}{\partial b_{j}^-} + sign(p_{k}^- - y_{nk}^-)\frac{\partial p_{k}^-}{\partial b_{j}^-}
\label{eq20_bias}
\end{equation}
where,

\begin{flalign*}
\frac{\partial p_{k}^+}{\partial b_{j}^-} &= & \\
\end{flalign*}
\begin{align}
\left(1-\frac{1}{\textit{A} _k}\right)\mathbbm{1}(z_{k}^+>z_{k})
\frac{1}{\textit{A} p_k}\mathbbm{1}(p_{j}^- + w_{jk}^- >z_{k}^+) \nonumber\\
\left(1-\frac{1}{\textit{A} _j}\right)\mathbbm{1}(z_{j}^->z_{j})
\frac{1}{\textit{A} n_j}\mathbbm{1}(b_{j}^- >z_{j}^-)\nonumber\\
\label{eq21_bias}
\end{align}
and
\begin{flalign*}
\frac{\partial p_{k}^-}{\partial b_{j}^-} &= & \\
\end{flalign*}
\begin{align}
\left(1-\frac{1}{\textit{A} _k}\right)\mathbbm{1}(z_{k}^->z_{k})
\frac{1}{\textit{A} n_k}\mathbbm{1}(p_{j}^- + w_{jk}^+ >z_{k}^-) \nonumber\\
\left(1-\frac{1}{\textit{A} _j}\right)\mathbbm{1}(z_{j}^->z_{j})
\frac{1}{\textit{A} n_j}\mathbbm{1}(b_{j}^- >z_{j}^-)\nonumber\\
\label{eq22_bias}
\end{align}

The weights and bias are updated using the obtained error gradients during each iteration as follows;
\begin{equation}
{}_{}w_{ij,(\tau)}^{+}= {}_{}w_{ij,(\tau-1)}^{+} - \epsilon \frac{\partial E}{\partial w_{ij,(\tau-1)}^+} \\
\label{wgtij1}
\end{equation}
\begin{equation}
{}_{}w_{ij,(\tau)}^{-}= {}_{}w_{ij,(\tau-1)}^{-} - \epsilon \frac{\partial E}{\partial w_{ij,(\tau-1)}^-} \\
\label{wgtij2}
\end{equation}
\begin{equation}
{}_{}w_{jk,(\tau)}^{+}= {}_{}w_{jk,(\tau-1)}^{+} - \epsilon \frac{\partial E}{\partial w_{jk,(\tau-1)}^+} \\
\label{wgtjk1}
\end{equation}
\begin{equation}
{}_{}w_{jk,(\tau)}^{-}= {}_{}w_{jk,(\tau-1)}^{-} - \epsilon \frac{\partial E}{\partial w_{jk,(\tau-1)}^-} \\
\label{wgtjk2}
\end{equation}
\begin{equation}
{}_{}b_{j,(\tau)}^{+}= {}_{}b_{j,(\tau-1)}^{+} - \epsilon \frac{\partial E}{\partial b_{j,(\tau-1)}^+} \\
\end{equation}
\begin{equation}
{}_{}b_{j,(\tau)}^{-}= {}_{}b_{j,(\tau-1)}^{-} - \epsilon \frac{\partial E}{\partial b_{j,(\tau-1)}^-} \\
\end{equation}
\begin{equation}
{}_{}b_{k,(\tau)}^{+}= {}_{}b_{k,(\tau-1)}^{+} - \epsilon \frac{\partial E}{\partial b_{k,(\tau-1)}^+} \\
\end{equation}
\begin{equation}
{}_{}b_{k,(\tau)}^{-}= {}_{}b_{k,(\tau-1)}^{-} - \epsilon \frac{\partial E}{\partial b_{k,(\tau-1)}^-} \\
\end{equation}
where $\epsilon$ is the learning rate and $\tau$ indicates the iteration step.
\section{Conventional MLP and MP-MLP training complexity: Proof}\label{appendixc}
\subsection{Conventional MLP}

Consider a three layer conventional MLP with I input nodes, J hidden nodes and $K=1$ output nodes as shown in \figref{fig_mlp}
\subsubsection{Forward Pass}
\subsubsection*{From layer I to layer J},
\begin{equation}
S_{JT}= W_{JI} \times X_{IT}
\label{eq_mlp_fw1}
\end{equation}
Here T is the total number of training samples, $W_{JI}$ is the weight matrix from layer J to I and $X_{IT}$ is the input sample matrix with T training samples.

Complexity for \eqref{eq_mlp_fw1} is,
\begin{equation}
C=T\times J\times I [C_A+ C_M]
\label{cmp_fw1}
\end{equation} 
$C_A$ and $C_M$ are complexity for addition and multiplication respectively.

Applying activation function $f$ to \eqref{eq_mlp_fw1},

\begin{equation}
Z_{JT}= f(S_{JT})
\label{eq_mlp_fw2}
\end{equation}

Complexity for \eqref{eq_mlp_fw2} is,
\begin{equation}
C=T\times J
\label{cmp_fw2}
\end{equation} 
\subsubsection*{From layer J to layer K},
\begin{equation}
S_{KT}= W_{KJ} \times Z_{JT}
\label{eq_mlp_fw3}
\end{equation}
where $W_{KJ}$ is the weight matrix from layer K to J 

The complexity of \eqref{eq_mlp_fw3} is
\begin{equation}
C=T\times J\times K [C_A+ C_M]
\label{cmp_fw3}
\end{equation} 

Applying activation function $f$,

\begin{equation}
Z_{KT}= f(S_{KT})
\label{eq_mlp_fw4}
\end{equation}

whose complexity is,
\begin{equation}
C=T\times K
\label{cmp_fw4}
\end{equation} 

\subsubsection{Back-Propagation}

\subsubsection*{Layer J K }

Error matrix at layer K is given as,

\begin{equation}
E_{KT}= f'(S_{KT}) \odot (Z_{KT}- O_{KT})
\label{eq_mlp_bp1}
\end{equation}
Here , $\odot$ indicates element-wise product and  $O_{KT}$ indicates the target matrix

The complexity of \eqref{eq_mlp_bp1} is,
\begin{equation}
C=T\times K [C_A+ C_M]
\label{cmp_bp1}
\end{equation} 

The delta matrix between layer J and K is,

\begin{equation}
D_{KJ}= E_{KT} \times Z_{TJ}
\label{eq_mlp_bp2}
\end{equation}
whose complexity is,
\begin{equation}
C=T\times K \times J [C_A+ C_M]
\label{cmp_bp2}
\end{equation}

The complexity of weight update step

\begin{equation}
W_{KJ}= W_{KJ} - D_{KJ}
\label{eq_mlp_bp3}
\end{equation}
is,
\begin{equation}
C= K \times J [C_A]
\label{cmp_bp3}
\end{equation}

\subsubsection*{Layer I J}

Error matrix at layer J is given as,

\begin{equation}
E_{JT}= f'(S_{JT}) \odot (W_{JK} \times E_{KT})
\label{eq_mlp_bp4}
\end{equation}

whose complexity is,
\begin{equation}
C=T\times K \times J [C_A+ C_M] + J\times T [C_M]
\label{cmp_bp4}
\end{equation} 

The complexity of delta matrix computation step

\begin{equation}
D_{JI}= E_{JT} \times X_{TI}
\label{eq_mlp_bp5}
\end{equation}

is,
\begin{equation}
C=T\times J \times I [C_A+ C_M]
\label{cmp_bp5}
\end{equation} 

Weights between layer I and J is updates as,

\begin{equation}
W_{JI}= W_{JI} - D_{JI}
\label{eq_mlp_bp6}
\end{equation}
whose complexity is,
\begin{equation}
C= I \times J [C_A]
\label{cmp_bp6}
\end{equation} 

From \cref{cmp_fw1,cmp_fw2,cmp_fw3,cmp_fw4,cmp_bp1,cmp_bp2,cmp_bp3,cmp_bp4,cmp_bp5,cmp_bp6}, the overall complexity is given as,

\begin{flalign}
C_{MLP-C}= \nonumber\\
& C_A[3JKT + 2JIT + JI + JK + KT]\nonumber\\
& + C_M[3JKT +2JIT  + JT + KT] + JT +KT
\label{cmp_bp7}
\end{flalign}

Since K=1 the overall complexity becomes,

\begin{flalign}
CT_{MLP-C}= \nonumber\\
& C_A[3JT + 2JIT + JI + J + T] \nonumber\\
&  + C_M[3JT +2JIT  + JT + T] + JT + T
\label{cmlp_finalappendix}
\end{flalign}

\subsection{MP MLP}
\subsubsection{Forward Pass}
Forward pass during training involves the same steps as that of inference, 
\subsubsection*{Layer I J }

The complexity of  \cref{eq14,eq15},  is given as,

\begin{equation}
C=  2 \times J\times T [2IC_A + F \times log(2I) C_c] 
\label{mpcmp_fp1}
\end{equation} 
Here $C_c$ is the complexity of comparison operation, T is the total number of training samples, J is the number of hidden neurons, I is the number of input neurons and K is the number of output neurons, which is 1 considering a 2-class problem. F is the sparsity factor.

\subsubsection*{Layer J K}

Similarly for \cref{eq3,eq4}, the complexity is given as,

\begin{equation}
C=  2 \times T [2JC_A + F \times log(2J) C_c] 
\label{mpcmp_fp2}
\end{equation} 
as $K=1$
\subsubsection{Back-propagation}

Similar to that of conventional MLP, we consider the weights update steps for the computation of complexity of back-propagation for MP MLP, assuming the same set of parameters as that of conventional MLP.
\subsubsection*{Layer JK}
The complexity of \cref{eq5,eq10} is given as,
\begin{equation}
C=  4 \times T \times C_A
\label{mpcmp_bp1}
\end{equation} 

\Cref{eq8,eq9,eq11,eq12} can be implemented by simple comparison and shift operations. 

Hence the total shift and comparison operations required to implement \cref{eq8,eq9,eq11,eq12} for $K=1$ are,

\begin{equation}
C=  \big[2\times T + 4\times T \times J\big] C_c + \big[4 \times T \times J\big] C_{S}
\label{mpcmp_bp2}
\end{equation} 
Here $C_c$ is the complexity of comparison operation and $C_{S}$ is the complexity of shift operation.

Now considering the weight update steps in \cref{wgtjk1,wgtjk2} for $W_{JK}$	for $K=1$ and, the complexity is given as,

\begin{equation}
C=   2 J \times C_A
\label{mpcmp_bp3}
\end{equation} 

\subsubsection{Layer IJ}
For the layer IJ considering \cref{eq16,eq20,eq17,eq19,eq21,eq22,wgtij1,wgtij2}, the total addition operations required is given as,

\begin{equation}
C=  \big[ 4 \times T + 4 \times T \times I \times J + 2 \times I \times J\big] C_A
\label{mpcmp_bp4}
\end{equation} 
For \cref{eq17,eq19,eq21,eq22}, the total unique comparison operations required (ignoring those computed in the previous steps) are,
\begin{equation}
C=  \big[ 2 \times T \times J + 4 \times T \times I \times J ] C_c
\label{mpcmp_bp5}
\end{equation} 

and the total unique shift operations required are,
\begin{equation}
C=  \big[ 8 \times T \times I \times J + 4 \times T \times J ] C_{S}
\label{mpcmp_bp6}
\end{equation}

Considering \cref{mpcmp_fp1,mpcmp_fp2,mpcmp_bp1,mpcmp_bp2,mpcmp_bp3,mpcmp_bp4,mpcmp_bp5,mpcmp_bp6}, the overall complexity is given as,

\begin{flalign}
CT_{MLP-MP}= \nonumber\\
& C_A\big[8JIT + 4JT + 8T + 2J + 2JI\big]   \nonumber\\
& +  C_c \big[2JT \times F \times log(2I)  + 2T \times F \times log(2J)  \nonumber\\
& + 4JIT + 6JT + 2T\big]\nonumber\\
& + C_{S}\big[8JT + 8JIT\big]
\end{flalign}

\end{document}